\documentclass[11pt]{article}

\usepackage[preprint]{acl}

\usepackage{times}
\usepackage{latexsym}

\usepackage[T1]{fontenc}

\usepackage[utf8]{inputenc}

\usepackage{microtype}

\usepackage[varqu,varl]{inconsolata}

\usepackage{graphicx}

\usepackage{times}
\usepackage{latexsym}
\usepackage[T1]{fontenc}
\usepackage[utf8]{inputenc}
\usepackage{microtype}
\usepackage{inconsolata}
\usepackage{graphicx}
\usepackage{booktabs}

\usepackage{amsmath}
\usepackage{amssymb} 
\usepackage{xcolor}
\usepackage{tcolorbox}        
\tcbuselibrary{skins,breakable}
\usepackage{pifont}
\usepackage{algorithm}
\usepackage{algpseudocode}
\usepackage{placeins}
\usepackage{xurl}

\newcommand{\revision}[1]{#1}
\newcommand{\prompttag}[1]{\textcolor{blue!70!black}{\texttt{#1}}}
\newcommand{\Description}[1]{}              

\definecolor{ScriptedBlue}{RGB}{0, 0, 255}
\definecolor{GeneratedOrange}{RGB}{255, 165, 0}
\definecolor{npcBlue}{HTML}{4F81BD}       
\definecolor{cohGreen}{HTML}{9BBB59}      
\definecolor{agencyOrange}{HTML}{F79646}  

\title{Towards Self-Evolving Narrative Systems: Real-Time Adaptation and Long-Term Optimization with LLMs}

\author{
    \textbf{Ruisheng Sun, Jiaqi Shao, Weijia Han, Linxi Ju, Yunzhen Zhang,} \\
    \textbf{Binyan Sun, Rong Chen, Ye Lu, Rui Hu, Bing Luo} \\
    Duke Kunshan University \& Duke University \\[0.5em]
    \textbf{Benjamin Blas Maristany, Ernesto Escobar} \\
    Duke University
}

\begin{document}
\maketitle
\begin{abstract}
Traditional interactive narrative systems offer strong authorial control and structural coherence but often restrict player agency. In contrast, LLM-driven narratives enable open-ended generation and greater flexibility, yet frequently suffer from inconsistency and diminished narrative control. To navigate this trade-off, we propose a hybrid narrative framework that dynamically balances pre-authored content with real-time LLM generation via semantic similarity alignment. Additionally, sustaining narrative quality over extended gameplay remains a core challenge in LLM-based storytelling. To address this, we introduce a self-evolving, feedback-driven optimization framework that leverages in-context learning to iteratively refine prompts, eliminating the need for manual prompt engineering. Through a case study and participant-based evaluation, we show that our framework enhances narrative coherence while preserving player agency, effectively aligning generated content with both authorial intent and evolving player behavior.
\end{abstract}

\section{Introduction}

\begin{figure*}[t!] 
    \centering 
    \includegraphics[
        width=0.9\textwidth,
        alt={Overview of the dynamic narrative generation system. It begins with pre-generated mode, where authored scripts form structured plots, progresses to gameplay mode, where the system either selects pre-written paths or generates new ones, and concludes with a feedback loop that refines prompts over time, making the system self-evolving.}
    ]{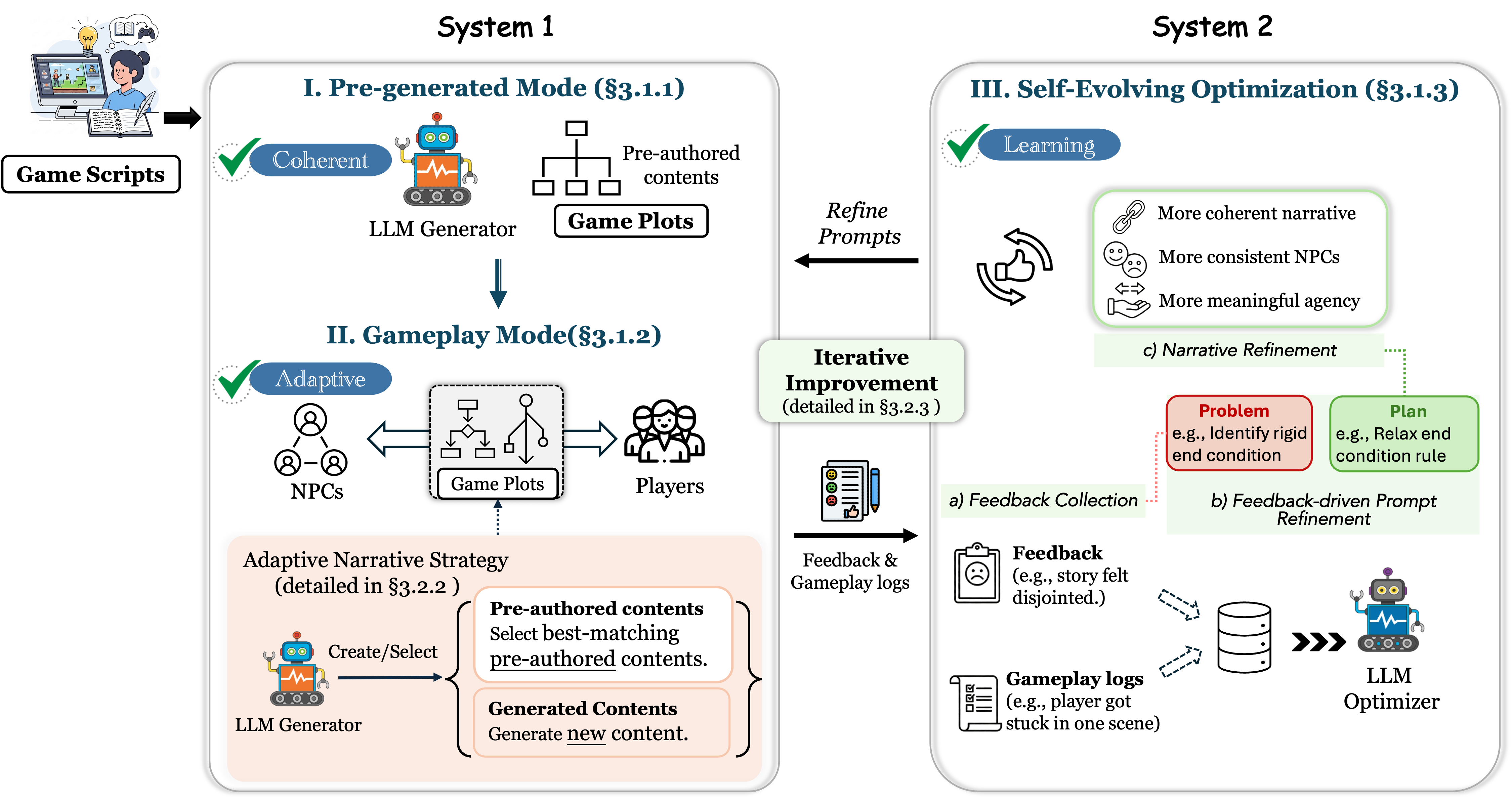} 
    \caption{
        \textbf{System Architecture Overview.} Our framework addresses coherence, adaptability, and continual improvement via three interdependent phases: \textbf{I}. \textit{Pre-generated Mode} transforms author-defined \textbf{Game Scripts} into modular \textbf{Game Plots} offline using an {LLM Generator}, ensuring narrative structure and quality; \textbf{II}. \textit{Gameplay Mode} dynamically selects between \textit{pre-authored} and \textit{generated content} using an \textbf{adaptive narrative strategy} that responds to player choices; and \textbf{III}.~\textit{Self-Evolving Optimization} uses player feedback to iteratively refine prompts via an {LLM Optimizer}, enhancing future generation. These refined prompts guide the {LLM Generator} to produce contextually grounded narratives that reflect both authorial design and player behavior.
    }
    \label{fig:full_system_overview}
\end{figure*}

For decades, designers of interactive narratives have pursued the dual goals of crafting compelling, coherent stories and of affording players agency ~\citep{domsch2013storyplaying, tanenbaum2010agency, 10.1145/3411764.3445540}: the ability to take meaningful actions that influence the game or story. 
Traditional systems based on branching storylines and fixed dialogue trees provide reliable and well-crafted experiences.
However, their reliance on predefined choices often results in predictable outcomes, limiting players' ability to shape the story and reducing the sense of personal agency.

Recent advances in Large Language Models (LLMs) offer new capabilities for interactive narrative systems by generating real-time text and responsive character behavior~\citep{andres2024assemblages}, enabling narratives that can dynamically adapt to player input. \revision{However, this flexibility introduces fundamental challenges: LLMs' inherent stochasticity and context sensitivity can lead to narrative incoherence, inconsistent character behavior, and loss of authorial control~\citep{10.1145/3706599.3719971,alavi2024gameplot}.} 

\revision{Prior work has explored various generation strategies, including constraining LLM outputs within predefined narrative structures~\cite{sun2023language, earle2025dreamgarden}, while others focus on personalization~\cite{villareale2023integrating,fu2025like} and leveraging multi-agent coordination for long-form generation~\cite{zhang2024storywriter} and PANGeA~\cite{buongiorno2024pangea}. However, these systems reveal \textbf{persistent trade-offs}: those prioritizing flexible generation often suffer from degraded coherence over time~\cite{peng2024player,jeong2025ligs}. \revision{More critically, they remain \textbf{fundamentally static} in their generation logic, i.e., their prompts and generation strategies are fixed once designed. The recent SCORE framework demonstrates that even well-engineered static prompts cannot maintain quality across extended interactions~\citep{yi2025score}, highlighting the need for a narrative system that adapts and improves over time.}}

To address these limitations, we propose a two-system framework that progressively builds narrative capabilities:

\revision{\textbf{System 1: Narrative Structure and Adaptive Strategy} adaptively decides between reusing pre-authored content and generating new narrative in real-time, based on how well player actions align with the intended story direction. We model the game plot as a \textit{hierarchical narrative tree} whose authored nodes form a reusable backbone, while runtime generation inserts context-constrained branches when player action deviate.  This adaptation is guided by \textit{semantic alignment score}~\cite{wang2020minilm} that quantifies the match between player actions and narrative context.

\textbf{System 2: Self-Evolving Optimization} enhances System 1 by refining the prompts used in its narrative generation, which strengthens long-term coherence and character consistency.
It analyzes player feedback across \textit{narrative coherence}, \textit{NPC consistency}, and \textit{player agency} via feedback-driven optimization with In-Context Learning (ICL~\cite{madaan2023selfrefine}), thereby diagnosing narrative issues and refining the prompts that guide System 1's generation, ultimately improving storytelling quality at scale without manual prompt engineering.

We evaluate our approach through a case study based on the ``Escape from Fort Joy'' narrative arc~\cite{EscapeFromFortJoyWiki} from \textit{Divinity: Original Sin 2}~\cite{DivinityOriginalSin2}, showing that our system maintains narrative coherence while giving players meaningful choices and automatically improves over time based on player feedback.}

To summarize, our contributions are:

\begin{enumerate}
    \item \textbf{Narrative Structure and Adaptive Strategy}: We introduce a modular narrative construction framework that transforms pre-authored scripts into reusable narrative graphs, together with a \textit{semantic alignment} mechanism that dynamically decides when to follow authored branches versus generating context-constrained new content.
    \item \textbf{LLM-Driven Self-Evolving Optimization}: We develop an automated feedback-to-improvement system in which our LLM-based Optimizer analyzes player feedback, diagnoses narrative issues, and generates refined prompts—creating a closed-loop optimization cycle for continuous quality improvement (aligning with continual human-in-the-loop optimization principles \citep{liao2025chilo}).
    
    \item \textbf{Human-Centered Evaluation and Impact}: Our user studies (N=12) demonstrate that the system evolves through long-term player interaction, balancing authorial control and generative freedom to improve coherence and immersion.
\end{enumerate}

We will release our code and the Fort Joy scenario specification to facilitate reproducibility.

\section{Related Work}
\label{sec:related_work}

\revision{\paragraph{\textbf{Balancing Narrative Coherence and Player Agency}}
Player agency-the player’s ability to take actions or make choices that meaningfully shape the narrative~\cite{tanenbaum2010agency,10.1145/3411764.3445540,cavazza2017interactive}—has long conflicted with narrative coherence.Traditional branching structures (e.g., Twine, Ink) restrict actions~\cite{domsch2013storyplaying}, while the GFI framework formalizes narrative via Goals, Feedback, and Interpretation but scales poorly to unanticipated inputs~\cite{cardona2020gfi}. Even modular frameworks like WhatELSE~\cite{ruan2024whatelse} require manual specification, fixing coherence at authoring time. Short's taxonomy~\cite{short2016beyond} further shows rigid branching fails for player-driven progression. In the ICIDS tradition, Goins~\citep{goins2018structuring} proposes structural manipulation of space and interactivity as a storytelling model, and Ferri~\citep{ferri2015idn} analyzes how narrative structures operate across IDN authoring systems. 
Together, these suggest pre-authored structures alone cannot ensure coherence when player intent is unpredictable; with LLMs, structural guarantees weaken, motivating continuous realignment of generative output during gameplay.}

\revision{\paragraph{\textbf{LLM‑Driven Story Generation}}
LLM assistance in narrative creation spans designer and player workflows~\cite{green2018authoring}. In designer-facing workflows, Su et al.~\citep{su2024game}, Wang et al.~\citep{wang2025aideation}, and Chen et al.~\citep{coexploreds2025} show that designers leverage LLMs to brainstorm plot directions but still struggle with controlling tone, pacing, and adherence to intended narrative themes; generated content becomes inconsistent across iterations unless designers intervene with explicit corrective feedback~\citep{wang2025can}.}

\revision{For players, Zhang et al.~\cite{zhang2025canaiprompt} find multimodal agents can guide actions but drift in longer interactions; Peng et al.~\citep{peng2024player} similarly observe that LLM-driven NPCs produce engaging moment-to-moment dialogue but fail to track longer-term goals or maintain stable characterization. Wu et al.~\cite{wu2025playwriting} propose Playwriting-guided Generation and Plot-based Reflection to improve structure and alignment, while Wang et al.~\cite{wang2025cluecart} help interpret fragmented narratives. Yet the SCORE framework~\citep{yi2025score} quantifies that models frequently violate previously established facts and causal dependencies even when supplied with long contexts—demonstrating that static prompt-based control is insufficient for sustaining coherent narratives over extended play.}

\section{System Design}
\label{sec:system_design}
This section details the technical framework of our dynamic narrative generation system, which integrates LLMs with structured, author-defined narrative elements and continuous player feedback, balancing emergent narrative with coherent, high-quality progression.

\subsection{System Architecture Overview}
\label{sec:overview}

\revision{As shown in Fig.~\ref{fig:full_system_overview}, our architecture combines the reliability of pre-authored content with the flexibility of real-time generation, while continuously learning from player interactions. The fundamental insight driving our work is that effective interactive storytelling requires \textbf{three distinct but interconnected capabilities}: (1) generating coherent, high-quality narrative content offline; (2) adapting this content dynamically during gameplay; and (3) learning from player interactions and improving over time. We implement these capabilities through three interdependent phases:

\begin{itemize}
    \item[I.] \textbf{Pre-generated Mode} addresses the \textit{coherence problem}: LLMs, while creative, can produce inconsistent or low-quality content when generating narratives without sufficient context. This phase transforms human-authored \textbf{Game Scripts}—which specify the environment, NPC traits, decision points, and outcomes—into modular \textbf{Game Plots} offline by the LLM Generator (\S\ref{sec:pre_generated_sub}), ensuring baseline narrative quality and structural coherence.
    
    \item[II.] \textbf{Gameplay Mode} addresses the \textit{adaptation problem}:  Static pre-generated content cannot respond to the full spectrum of player actions. We propose an adaptive strategy that dynamically steers narrative progression through content selection and real-time generation (\S\ref{sec:adaptive_narrative_strategy}), using semantic alignment to determine when to reuse pre-generated content versus generating new narratives. 
     
    \item[III.] \textbf{Self-Evolving Optimization} addresses the \textit{learning problem}: Traditional game development relies on manual iteration cycles that are slow and difficult to scale. Our system integrates an LLM Optimizer that continuously: a) collects and analyzes individual player feedback and actions after each gameplay, b) concludes feedback features for improvements, c) refines prompts for the LLM Generator, and d) repeats as a loop (\S\ref{sec:con-opt}).
\end{itemize}}

\revision{The three phases operate as an integrated system: the pre-generated mode establishes the narrative foundation; the gameplay mode uses this foundation while collecting player data; and the self-evolving optimization refines the prompts used in the pre-generated mode, forming a continuous improvement loop.

In our implementation, the LLM Generator leverages \texttt{gpt-4o-mini}~\cite{OpenAI_GPT4o_mini_2024}  to generate and adapt narrative content in both pre-generated and gameplay modes, responding dynamically to both authorial design goals and player interactions. To ensure continued alignment between generated content, authorial intent, and evolving player behavior, we introduce a self-evolving optimization loop powered by an LLM Optimizer, instantiated by \texttt{gemini-2.5-flash}~\cite{Gemini2.5_Flash_2025}\footnote{See Appendix~\ref{app:model-selection} for the model selection rationale.}, which iteratively refines prompts and narrative strategies based on feedback.}

\subsection{Core Mechanisms and Implementation}
\label{sec:core_design}

\subsubsection{Pre-generated Mode}
\label{sec:pre_generated_sub}

The pre-generated mode constructs a coherent narrative space offline, balancing LLM creativity with \textbf{quality control and narrative consistency}. Traditional approaches face a fundamental trade-off: highly structured content (e.g., branching dialogue trees~\cite{akoury2023framework}) provides coherence but limits player agency, while unstructured generation (e.g., pure LLM improvisation) offers flexibility but risks narrative inconsistency and quality degradation~\cite{yi2025score}. Our solution is based on \textbf{hierarchical narrative decomposition}: breaking down complex narratives into modular, reusable components that can be dynamically composed and adapted while maintaining overall coherence.

The process begins with \textbf{human-authored} \texttt{Game Scripts}, which capture the author's design vision: defining environments, initializing NPCs, specifying end conditions, and formulating \textit{key questions} that mark pivotal decision points. These key questions serve as narrative anchors, focusing generation on \textit{specific transitions} rather than unconstrained text. From each key question, the system generates multiple \textbf{narrative paths}—each addressing the key question from different perspectives or character actions—and each path is realized as a sequence of modular \textbf{nodes}, the fundamental units of story execution. Each node encapsulates a distinct moment in the game world, containing its environmental state, active NPCs, triggering conditions, and the guiding key question. \revision{Inspired by the GFI framework~\cite{cardona2020gfi} and MoPS~\cite{ma2024mops}, which formalize narrative design and modular premise synthesis, this structure enables flexible reuse and dynamic selection for coherent, adaptive player experiences.}

The system generates paths and nodes through three complementary prompts (see Appendix~\ref{app:prompts} for full templates): a \prompttag{System Prompt} defines global rules, world constraints, and narrative guidelines; a \prompttag{Scripted Prompt} realizes the author's intended storyline to form \textbf{\textcolor{ScriptedBlue}{Scripted Branches}} (the canonical backbone); and a \prompttag{Generated Prompt} produces \textbf{\textcolor{GeneratedOrange}{Pre-generated Branches}}—LLM-generated alternatives that expand narrative flexibility while maintaining character consistency and thematic alignment. Together, these prompts populate each node with both branch types, allowing the system to select between them at runtime based on player input. See Appendix~\ref{app:narrative-generation-details} for the full architecture diagram (Fig.~\ref{fig:pre_generated_mode}).

\subsubsection{Adaptive Narrative Strategy}
\label{sec:adaptive_narrative_strategy}

During gameplay, the adaptive narrative strategy evaluates \textit{semantic alignment} between player input and in-game context to determine the most appropriate narrative path. \revision{Our approach draws on insights from~\cite{sreenarayanan2024emotion}, which emphasizes tailoring narrative progression to players' evolving behaviors.} Let $\mathbf{p} \in \mathbb{R}^d$ denote the embedding of the player's input and $\mathbf{n}_i \in \mathbb{R}^d$ the embedding of the $i$-th pre-generated node, both generated using MiniLM-L6-v2~\cite{wang2020minilm}. The alignment score is computed via cosine similarity:
$$
S(\mathbf{p}, \mathbf{n}_i) = \frac{\mathbf{p} \cdot \mathbf{n}_i}{\|\mathbf{p}\| \, \|\mathbf{n}_i\|},
$$
with $S \in [-1,1]$, where $1$ indicates perfect alignment and $-1$ complete opposition. Based on this score, the system applies three decisions:
\begin{enumerate}
\item \textbf{Freeform Exploration:} If $S(\mathbf{p}, \mathbf{n}_i) < \theta_{\text{low}}$, the system allows continued player exploration within the current node.
\item \textbf{Authored Progression:} If $S(\mathbf{p}, \mathbf{n}_i) \geq \theta_{\text{high}}$, the system selects the best-matching pre-generated node $j = \arg\max_i S(\mathbf{p}, \mathbf{n}_i)$.
\item \textbf{Emergent Adaptation:} If $\theta_{\text{low}} \leq S(\mathbf{p}, \mathbf{n}_i) < \theta_{\text{high}}$, the system generates a new node in real-time, considering scene state, NPC traits, and unresolved story elements.
\end{enumerate}
We set $\theta_{\text{high}} = 0.70$ and $\theta_{\text{low}} = 0.40$ (\S\ref{sec:experimental_setup}) to ensure pre-generated narrative selection when input closely aligns with the expected storyline, while allowing dynamic generation under player-driven interactions. The full procedure is summarized in Algorithm~\ref{alg:semantic-alignment} (Appendix~\ref{app:semantic-alignment-algorithm}), and Fig.~\ref{fig:Node Selection Strategy} (Appendix~\ref{app:adaptive-strategy-figure}) illustrates this strategy with a branching example.

\subsubsection{Self-Evolving Optimization via Prompt Refinement}
\label{sec:con-opt}

While the pre-generated mode and adaptive strategy address coherence and adaptation within a single playthrough, a critical limitation remains: once prompts are engineered, they remain \textbf{static}. As our evaluation shows, even well-designed prompts fail to cover all player behaviors—coherence degrades, character motivations become underspecified, and end conditions trigger inconsistently. Manual prompt engineering cannot scale to address every failure mode observed in extended play.

\begin{figure*}[t!]
    \centering
    \includegraphics[width=0.9\textwidth]{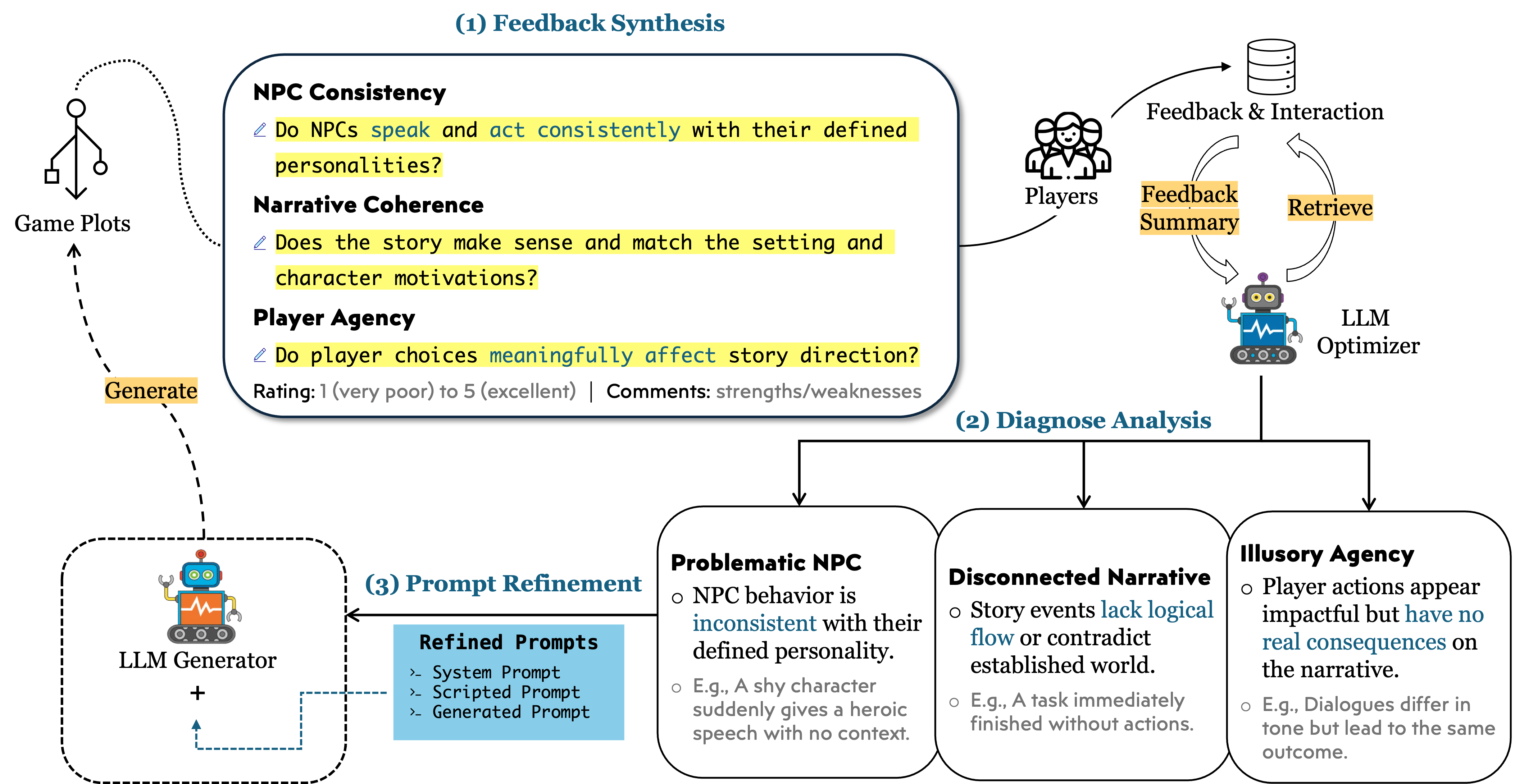}
\caption{
\textbf{Feedback-driven prompt refinement.}
(1) \textit{Feedback Synthesis}: Players provide ratings (1--5) and qualitative comments on \textit{NPC Consistency}, \textit{Narrative Coherence}, and \textit{Player Agency}, stored alongside interaction history.
(2) \textit{Diagnose Analysis}: The LLM Optimizer retrieves this data to identify issues such as inconsistent NPC behavior, disconnected narrative flow, or illusory agency.
(3) \textit{Prompt Refinement}: Refined system, scripted, or generated prompts are produced to resolve diagnosed issues and improve future narrative generation.
}
\label{fig:LLM-based Optimization Pipeline}
\Description{LLM-based Optimization Pipeline. Player feedback is collected, analyzed by an LLM-based optimizer to identify issues, and transformed into refined prompts. These prompts are reintegrated into the system, improving narrative coherence and NPC behavior across iterations.}
\end{figure*}

To overcome this, we introduce \textbf{self-evolving optimization}, an automated feedback loop that continuously refines the \texttt{System}, \texttt{Scripted}, and \texttt{Generated} prompts based on aggregated player feedback. As illustrated in Fig.~\ref{fig:LLM-based Optimization Pipeline}, this mechanism operates in four iterative steps:

\begin{enumerate}
    \item \textbf{Feedback Collection:} After each playthrough, the system captures quantitative ratings (on \textit{narrative coherence}, \textit{NPC consistency}, \textit{player agency}) and qualitative comments, along with interaction logs.
    \item \textbf{Diagnostic Analysis:} The \texttt{LLM Optimizer} aggregates feedback and identifies recurring issues. E.g.``NPC motivation underspecified in the \texttt{Generated Prompt}'' or ``end conditions too rigid to recognize natural player agreement.''
    \item \textbf{Targeted Prompt Refinement:} For each diagnosed issue, the \texttt{LLM Optimizer} proposes a concrete textual patch to the relevant prompt (see Appendix~\ref{app:optimization-example} for a worked example with NPC Fane).
    \item \textbf{Redeployment and Iteration:} The refined prompts are redeployed to both pre-generation and gameplay modes, creating a closed-loop optimization across rounds of play.
\end{enumerate}

\revision{Our optimization leverages In-Context Learning and self-reflective mechanisms~\cite{madaan2023selfrefine}: the \texttt{LLM Optimizer} evaluates its own narrative outputs, derives insights from player engagement, and stores reflections in memory, which inform subsequent story generation.} Algorithm~\ref{alg:prompt-refinement} summarizes this process.

\begin{algorithm}[t]
\caption{Feedback-Driven Prompt Refinement}
\label{alg:prompt-refinement}
\begin{algorithmic}[1]
\Require Player feedback $F$, interaction logs $L$, prompts $P = \{P_{\text{sys}}, P_{\text{scr}}, P_{\text{gen}}\}$
\Ensure Refined prompts $P'$
\State $D \gets \textsc{Diagnose}(F, L)$ \Comment{LLM Optimizer identifies issues}
\State $H \gets \textsc{GenerateHypothesis}(D)$ \Comment{e.g., ``NPC motivation underspecified''}
\For{each prompt $P_i \in P$}
    \State $\Delta \gets \textsc{ProposeModification}(H, P_i)$ \Comment{specific textual change}
    \State $P'_i \gets \textsc{ApplyPatch}(P_i, \Delta)$
\EndFor
\State \textsc{Deploy}$(P')$ \Comment{to pre-generation and gameplay modes}
\State \Return $P'$
\end{algorithmic}
\end{algorithm}

\revision{
\section{Experimental Setup}
\label{sec:experimental_setup}

We conducted a controlled user study to evaluate the two systems introduced in \S\ref{sec:system_design}. Because System 2 builds on System 1, we designed two \textit{sequential} tracks with the same 12 participants: \textbf{Track A} isolates System 1 by comparing our \textbf{structured Game Plot system} against a \textbf{flat baseline} that generates entire scenes in a single LLM call without modular structure or end-condition constraints; \textbf{Track B} then isolates System 2 by comparing the \textbf{self-evolving (optimized)} version of System 1 against its \textbf{static (unoptimized)} counterpart. Thus, Track A's structured system serves as Track B's static baseline, and both tracks used a within-subject design with counterbalanced condition order.

\subsection{Participants}
We recruited 12 participants (age 18--36) from our university community, comprising experienced RPG players (n=10) and professional game designers (n=2); see Appendix~\ref{app:irb-constraints} for details. Each participant completed both tracks in a single session of approximately 240 minutes, with a one-day break between the two tracks to minimize carryover effects. Within each track, participants experienced both conditions in counterbalanced order. With a within-subject design across two tracks, N=12 provides sufficient statistical power for detecting medium-to-large effect sizes (Cohen's $d \geq 0.8$) at $\alpha=0.05$. The study received ethics approval from an institutional review board, and all participants provided informed consent.

\subsection{Evaluation Dimensions}
Our evaluation targeted both \textit{story-level} and \textit{interaction} qualities: \textbf{narrative coherence} (global logical flow, absence of contradictions), \textbf{NPC consistency} (persona/voice stability and memory across scenes), and \textbf{player agency} (perceived influence over narrative direction and outcomes)~\citep{harrell2009agency, Poivet2025NPCConsistency}. These dimensions jointly assessed whether structure preserves coherence without sacrificing agency, and whether optimization increases believability and resilience over time. After each playthrough, participants completed a survey combining 5-point Likert scales and free-text responses~\citep{10.1145/1753326.1753686}.\footnote{To ensure participants focused on their experience rather than technical implementation, we avoided terms related to system architecture (e.g., ``key questions,'' ``node generation'').}

\subsection{Scene Data}
We authored a compact narrative arc of \textbf{nine story scenes} spanning the ``Escape from Fort Joy'' journey. Each scene was specified as a modular node with a location, trigger conditions, interactive NPCs, a set of \textit{key questions} (k1--k3) anchoring decision points, and a precise \textit{end condition} defining when the scene closes. Authorial structure set the \emph{order} and ending rules, while free-form input and node selection allowed \emph{how} each step unfolded to vary; key choices propagated across scenes (e.g., breaking Ifan's trust reduced support at Scenes~5 and~7). The full scene list, summaries, and key questions are provided in Appendix~\ref{app:scenes} (Table~\ref{tab:scene-overview}).

\subsection{Study Procedure}
Each participant received a standardized 10-minute orientation, including a 5-minute free-form tutorial with a non-story-critical NPC in a separate environment. In each track, participants experienced both conditions in counterbalanced order. For Track~A, participants experienced both \emph{Structured} and \emph{Flat} variants; for Track~B, both \emph{Optimized} and \emph{Static} variants. After each condition, they completed a structured survey assessing the key dimensions plus secondary measures (immersion, robustness to unexpected input, willingness to replay), concluding with open-ended questions. We systematically logged all player inputs, system responses, node selection decisions, end-condition triggers, and survey responses.

\subsection{Data Analysis}
\label{sec:data_analysis}
Following an inductive thematic-analysis procedure~\citep{braun2006using}, two researchers independently performed open coding on survey transcripts and in-game logs, resolved disagreements through discussion, and iteratively refined provisional themes through cross-referencing (e.g., aligning \textit{Opaque Endings} complaints with log segments showing repeated end-condition checks). Final themes were operationalized, linked to our quantitative metrics, and illustrated with representative quotations. The complete thematic map and full coding procedures are provided in Appendix~\ref{app:coding-process}.
} 

\revision{
\section{Results}
\label{sec:results}

Our evaluation yields insights into both the structural benefits of our narrative architecture and its capacity for self-improvement. We present findings from the two experimental tracks: Track A examines how author-guided hierarchical structuring balances coherence with player agency, and Track B investigates whether optimization improves narrative quality across multiple playthroughs.

\subsection{Track A: Evaluating Narrative Structuring for Coherence--Agency Balance}

Track~A evaluates System 1 by having every participant experience both the structured and flat systems in counterbalanced order, allowing us to attribute observed differences to narrative structure alone.

\begin{figure}[t]
    \centering
    \includegraphics[width=0.9\linewidth]{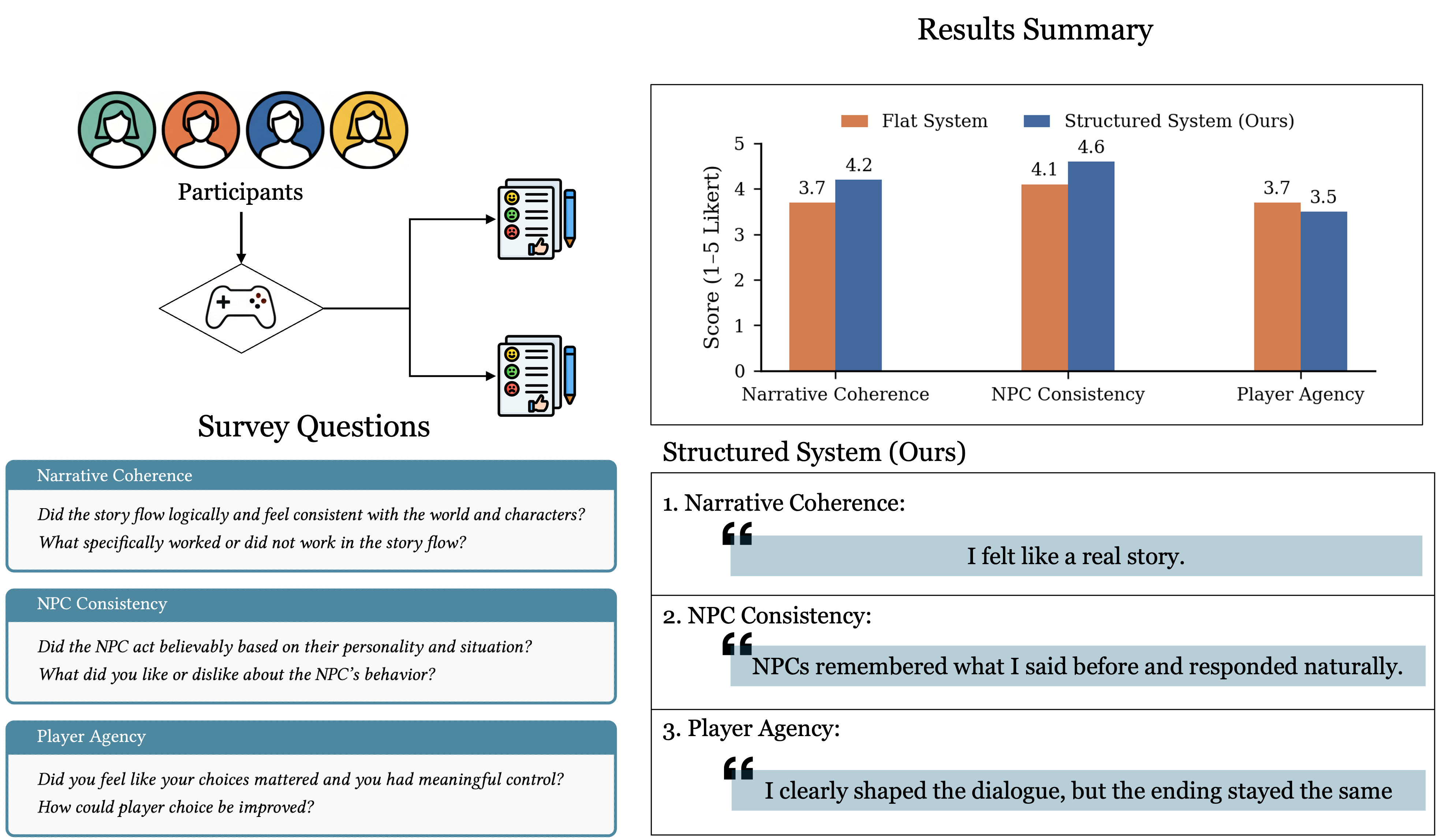}
\caption{\textbf{Track A Survey Results.} Compared to the flat system, our structured system improves narrative coherence (+0.5) and NPC consistency (+0.5), with only a modest reduction in player agency (--0.2).}
\vspace{-1em}
\label{fig:trackA}
\Description{Bar graph of Track A survey results comparing flat generation to narrative layering. Likert scale ratings show improvements in coherence (+0.5) and NPC consistency (+0.4) with only a modest reduction in player agency (-0.2) for the structured system.}
\end{figure}

\begin{table}[t]
    \centering
    \caption{Additional Track~A metrics (5-point Likert). }
    \label{tab:trackA}
    \resizebox{\columnwidth}{!}{%
    \begin{tabular}{lcc}
    \toprule
    \textbf{Metric} & \textbf{Flat System} & \textbf{Structured System} \\
    \midrule
    Immersion                  & 4.0 & 4.3~\textcolor{red}{($\uparrow$0.3)} \\
    Handling unexpected inputs & 3.6 & 4.1~\textcolor{red}{($\uparrow$0.5)} \\
    Willingness to replay      & 3.4 & 3.8~\textcolor{red}{($\uparrow$0.4)} \\
    \bottomrule
    \end{tabular}%
    }
\end{table}

Fig.~\ref{fig:trackA} shows that the structured system improves \emph{narrative coherence} (\textcolor{cohGreen}{+0.5}) and \emph{NPC consistency} (\textcolor{npcBlue}{+0.5}) with only a modest decrease in \emph{player agency} (\textcolor{agencyOrange}{--0.2}), indicating a typical coherence--agency trade-off.

To understand why, we examine thematic feedback. The structured system reduces \textit{Memory Failures} (\textcolor{npcBlue}{9/12 $\to$ 2/12}) and increases \textit{Consistent Persona} (\textcolor{npcBlue}{2/12 $\to$ 11/12}), while also reducing \textit{Opaque Endings} (\textcolor{cohGreen}{7/12 $\to$ 3/12}) and increasing \textit{Earned Closure} (\textcolor{cohGreen}{1/12 $\to$ 9/12}). For example, one participant noted in the flat system: \textit{``There were definitely moments of inconsistency... After I killed an enemy, I entered `he didn't die, just injured,' Ifan believed it''} (P11), whereas under the structured system: \textit{``When the Red Prince adapted to my unexpected input of confessing my affection... it was clear that my input matters''} (P11).

While the structured system improves coherence, it also increases \textit{Choice Stagnation} (\textcolor{agencyOrange}{1/12 $\to$ 4/12}), as one participant noted: \textit{``It felt like I had options, but no matter what I picked the story always returned to the same ending''} (P7). Yet it enhances \textit{Semantic Adaptation} (\textcolor{agencyOrange}{1/12 $\to$ 9/12}), reflecting stronger uptake of unusual inputs. Thus, stronger structure limits certain action-level freedoms but fosters a \emph{relational agency} grounded in responsiveness to player intentions. Table~\ref{tab:trackA} further shows improvements in \emph{immersion} (+0.3), \emph{handling of unexpected inputs} (+0.5), and \emph{willingness to replay} (+0.4).

\begin{tcolorbox}[colback=gray!5!white, colframe=ScriptedBlue!75!white, title=Track A: Takeaway]
Author-guided hierarchical structuring yields \textbf{coherent stories} and maintains \textbf{character consistency}, while preserving meaningful agency. Flat generation can be inventive but risks opaque endings and memory failures; structure mitigates these failures with only a minor agency trade-off.
\end{tcolorbox}

\subsection{Track B: Evaluating Optimization for Long-Term Evolution}

Track~B evaluates System 2 by investigating whether feedback-driven optimization improves System 1's narrative quality across playthroughs. Optimization was applied in \textit{rounds}: after each round, the LLM Optimizer processed aggregated feedback and logs to refine prompts, which were deployed for the next round.

\begin{figure}[t]
    \centering
    \includegraphics[width=0.9\linewidth]{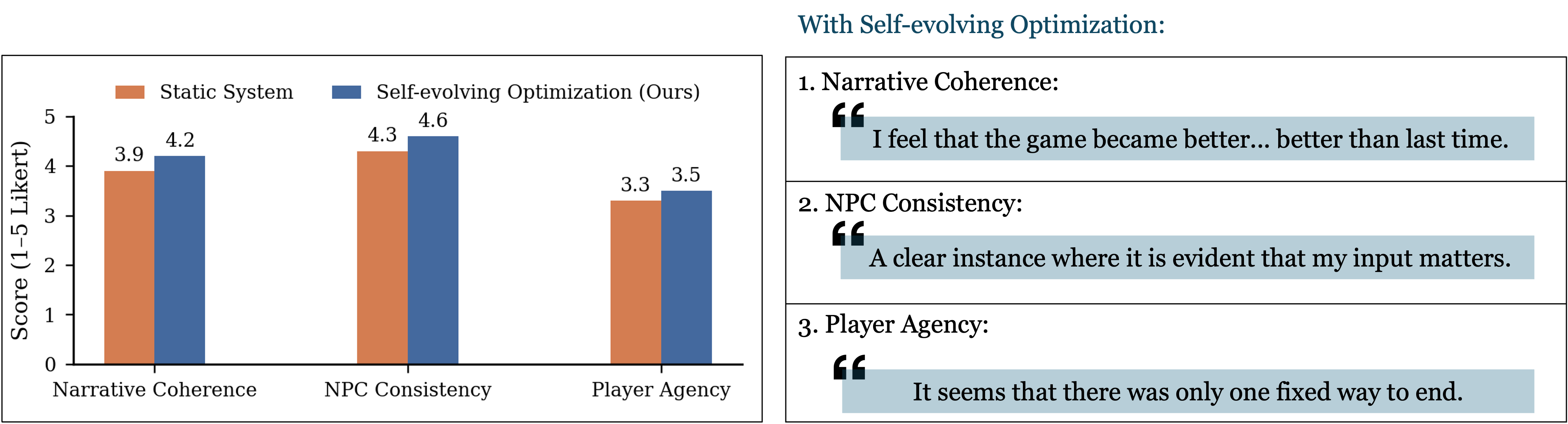}
    \caption{
\textbf{Self-evolving optimization vs.\ static baseline} (5-point Likert). Iterative refinement improves coherence and NPC consistency by +0.3 each and agency by +0.2. Gains accumulate across cycles.  
\textit{Takeaway}: Lightweight player-feedback loops tighten character voice and narrative logic without retraining.
}
    \label{fig:trackB}
    \Description{Track B survey results showing iterative optimization benefits comparing self-evolving optimization vs. static baseline systems. Narrative coherence (4.6 vs. 4.3), NPC consistency (4.6 vs. 4.3), and player agency (3.5 vs. 3.3) improved over three refinement cycles.}
\end{figure}

Fig.~\ref{fig:trackB} shows that the optimized system improves \emph{narrative coherence} (\textcolor{cohGreen}{+0.3}), \emph{NPC consistency} (\textcolor{npcBlue}{+0.3}), and \emph{player agency} (\textcolor{agencyOrange}{+0.2}). Statistical analysis (n=12$\times$3, within-subjects) revealed significant improvements: coherence 3.9$\to$4.2 ($p<0.05$), NPC consistency 4.3$\to$4.6 ($p<0.05$), and agency 3.3$\to$3.5 ($p<0.1$). These consistent gains suggest our optimization addresses multiple aspects simultaneously. The optimization trend across rounds is illustrated in Appendix Fig.~\ref{fig:eval_multiround_overview}.

To understand how optimization achieves these improvements, we examine thematic feedback. Iterative optimization increased \textit{Consistent Persona} (\textcolor{npcBlue}{6/12 $\to$ 11/12}), reduced \textit{Opaque Endings} (\textcolor{cohGreen}{7/12 $\to$ 3/12; $\downarrow$57.1\%}), and increased \textit{Earned Closure} (\textcolor{cohGreen}{1/12 $\to$ 9/12}). Participants described NPCs as ``alive, sharp, and consistent in tone,'' recalling moments where NPCs remembered earlier choices. For example: \textit{``Fane felt like a real schemer... It felt like I was actually negotiating with Fane''} (P5); and \textit{``characters felt like they had actual memories and personalities that persisted''} (P5).

While optimization improves coherence, Scene~7 still looped for non-canonical companion choices (P7, P11), indicating that end-condition coverage needs expansion---structural limitations require architectural improvements beyond prompt refinement. Nevertheless, \textit{Semantic Adaptation} increased (\textcolor{agencyOrange}{1 $\to$ 9}) and \textit{Choice Stagnation} dropped (\textcolor{agencyOrange}{8/12 $\to$ 2/12}), aligning with the +0.2 agency gain ($p<0.1$). Thus, optimization also fosters a more authentic agency grounded in responsiveness to player intentions.

\begin{tcolorbox}[colback=gray!5!white, colframe=ScriptedBlue!75!white, title=Track B: Takeaway]
Our system with self-evolving optimization continuously improves narrative quality. It reduces incoherence, strengthens character identity, and accumulates gains across rounds. Combined with Track A, it shows how our narrative system can evolve through long-term player interaction, achieving both \textbf{narrative coherence} and \textbf{NPC consistency}.
\end{tcolorbox}

\paragraph{Example: Eliminating ``Opaque Endings'' via Optimization.}
System 2 resolves \textit{Opaque Endings} by refining prompt logic to recognize semantic intent rather than exact phrases (e.g., ``I accept your terms''). In Scene~1, this reduced opaque endings from 7/12 to 3/12 ($\downarrow$57.1\%) and increased earned closure from 1/12 to 9/12. For example, ``I'm in, let's escape together'' now triggers the end condition. See Appendix~\ref{app:opaque-endings-workflow} for the full workflow and comparison.

\subsection{Discussion}
\label{sec:discussion}

Our study shows that a hybrid architecture combining authorial structure with feedback-driven optimization effectively manages the \textit{coherence--agency trade-off}. System~1 improved coherence (+0.5) and NPC consistency (+0.5) with only modest agency reduction. Critically, this trade-off is dynamic: Track~B shows that System~2's optimization mitigates agency constraints, reducing \textit{Choice Stagnation} (8/12 $\to$ 2/12) and increasing \textit{Earned Closure} across rounds. Rather than zero-sum, coherence and agency can be reconciled through \textit{Emergent Adaptation} (semantic routing) and character-consistent responses in \textit{Freeform Exploration}.

Our framework addresses limitations of both static authoring tools (e.g., \textit{Ink}\footnote{\url{https://www.inklestudios.com/ink/}}, \textit{Twine}\footnote{\url{https://twinery.org/}}) and fully generative systems (e.g., \textit{AI Dungeon}\footnote{\url{https://aidungeon.com/}}): the former cannot adapt to unforeseen player behaviors during play, while the latter suffer from narrative drift and inconsistency in extended sessions~\cite{yi2025score}. These challenges highlight the need for systems that go beyond single-pass generation and instead support \textit{continuous learning} and \textit{player-informed refinement}.
}

\section{Conclusion}

We introduced a self-evolving, LLM-driven narrative system for runtime adaptation and continual refinement with player engagement. Our system negotiates between pre-authored and generative content through semantic alignment, sustaining both coherence and freedom. Empirical results show that our approach improves coherence and preserves player agency under long-term play. Our work shifts the focus from one-off narrative generation to \textit{adaptive, player-informed systems} that evolve over time, pointing toward a new direction for sustainable, intelligent narrative system.

\section*{Limitations}
Our semantic alignment mechanism has three limitations. First, cosine similarity captures lexical overlap but not the speaker's intended meaning, so it cannot distinguish refusal from postponement or ironic agreement when embeddings are similar; this contributes to residual opaque endings (3/12 in the final optimization round). Second, alignment thresholds were set heuristically from pilot observations rather than a principled optimization procedure; different thresholds would shift the balance between authored progression and emergent adaptation. Third, embedding similarity is less interpretable than symbolic intent classification, complicating audits of narrative decisions. Dedicated intent classification modules—possibly with structured knowledge representations—would decouple narrative closure from surface lexical similarity and improve robustness and explainability.

\paragraph{Sample and Generalizability.}
Our evaluation is based on a pilot study with 12 participants recruited from a university community, primarily experienced RPG players and game designers. While the within-subject design provides sufficient statistical power for detecting medium-to-large effect sizes, the small sample limits generalizability. We did not collect gender or educational background due to IRB data minimization requirements. Furthermore, our study focuses on a single narrative arc (``Escape from Fort Joy'') from one genre (CRPG-style text-driven narrative). 

\paragraph{System Constraints.}
We use two commercial LLMs via API (GPT-4o Mini for generation, Gemini 2.5 Flash for optimization). This setup imposes several constraints. Context-window limits restrict the narrative history that can be maintained; summarization may lose fine-grained details needed for long-term coherence. API latency ($\approx$1.4s for GPT-4o Mini) can disrupt fast-paced interaction. External APIs also raise reproducibility concerns, as model updates or deprecations may affect behavior across runs. Future work could explore locally fine-tuned LLMs, hierarchical context management, and persistent long-term memory.

\paragraph{Evaluation Methodology.}
Our mixed-methods evaluation combines Likert ratings with qualitative thematic analysis, which provides rich insights but also has limitations. The thematic analysis involved two researchers who resolved disagreements through discussion; formal inter-rater reliability was not computed prior to reconciliation. Mapping qualitative themes to quantitative metrics (e.g., aligning ``Opaque Endings'' complaints with end-condition logs) was interpretive rather than algorithmic. Our evaluation also relies on participant self-report; expert-based assessment would complement it and help gauge applicability in real game design and production. Finally, the optimized system was deployed in rounds rather than in real time during a single session, which prevents intra-session inconsistency but cannot capture continuous adaptation dynamics.

\section*{Acknowledgements}
This work is supported by the Duke Provost Fund.

\bibliography{sample-base}

\appendix
\section{Narrative Generation Architecture Details}
\label{app:narrative-generation-details}

\subsection{Three-Prompt Architecture for Path and Node Generation}
\label{app:three-prompt-details}

The system generates narrative paths and nodes through a three-prompt architecture that creates a hierarchical branching structure. Each key question from the Game Script is processed using three complementary prompts (see Appendix~\ref{app:prompts} for full templates):

\begin{itemize}
    \item \textbf{\prompttag{System Prompt}}: Defines global rules, world constraints, and narrative guidelines.
    \item \textbf{\prompttag{Scripted Prompt}}: Realizes the author's intended storyline with specific character arcs and plot points.
    \item \textbf{\prompttag{Generated Prompt}}: Enables adaptive generation of alternative narratives for unexpected player choices.
\end{itemize}

These prompts work together to create two distinct types of narrative branches within each node:

\begin{itemize}
    \item \textbf{\textcolor{ScriptedBlue}{Scripted Branches}}: Pre-authored progression that forms the canonical storyline backbone, ensuring coherence and consistency.
    \item \textbf{\textcolor{GeneratedOrange}{Pre-generated Branches}}: LLM-generated alternatives that expand narrative flexibility while maintaining character consistency and thematic alignment.
\end{itemize}

By structuring content around modular nodes and hierarchical branching, the system pre-generates coherent yet flexible narrative content that enables dynamic adaptation during gameplay while preserving story integrity.

\begin{figure*}[t]
    \centering
    \includegraphics[width=0.9\textwidth]{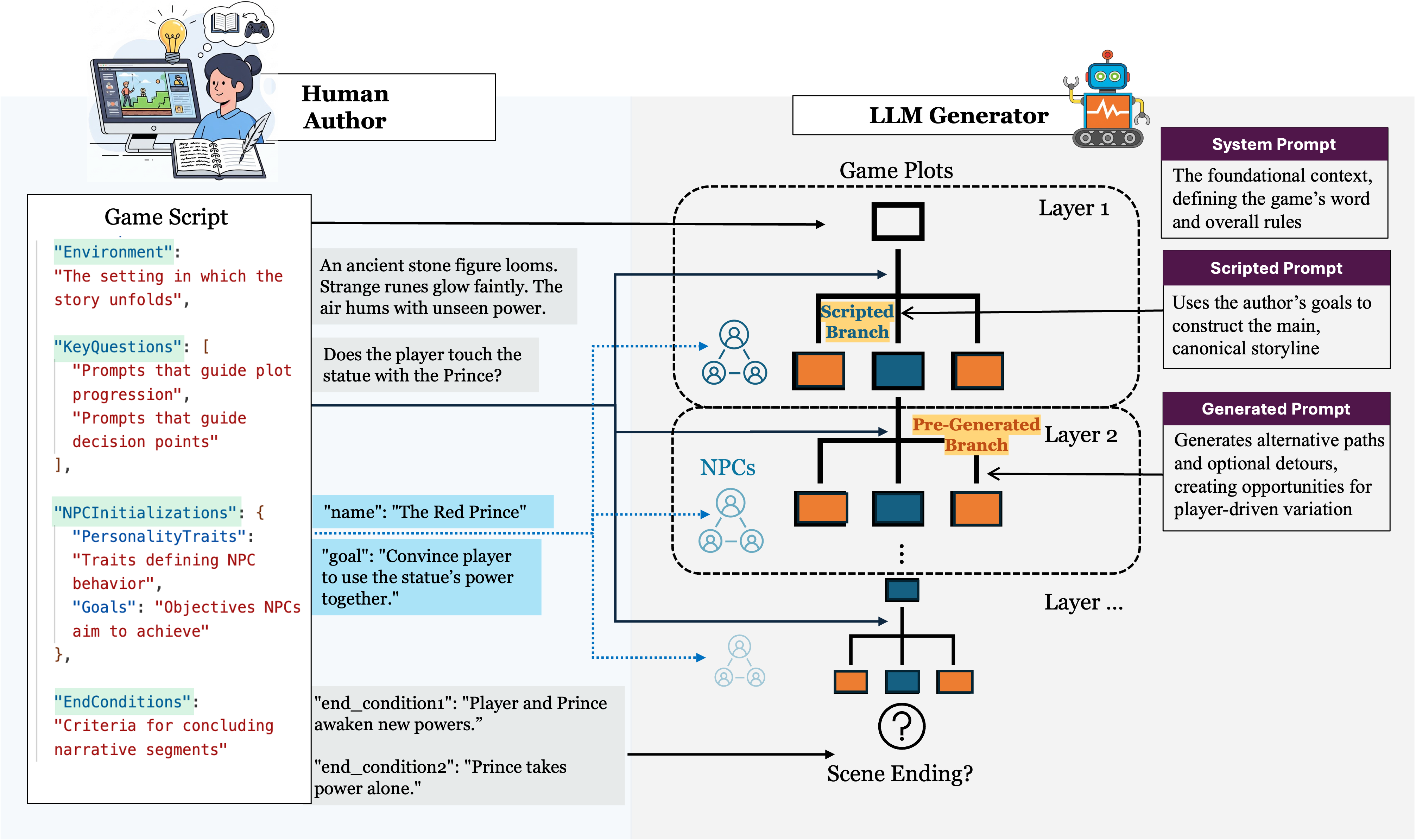}
\caption{
\textbf{Narrative Architecture Overview.} In the \textit{pre-generated mode}, authors design a \textbf{Game Script} that specifies the environment, NPC traits and goals, key questions, and ending conditions. This script is used to construct a \textbf{Game Plot}—a modular, hierarchical narrative structure comprising two types of branches: \textcolor{ScriptedBlue}{\textbf{Scripted Branches}}, realized by LLM Generator using \prompttag{Scripted Prompt}, follow the author's scripts. To support player-driven variability, the LLM Generator dynamically expands \textcolor{GeneratedOrange}{\textbf{Pre-generated Branches}} in response to decision points using \prompttag{Generated Prompt}. A global \prompttag{System Prompt} governs overarching narrative rules and constraints.
}
\label{fig:pre_generated_mode}
\Description{Diagram of the pre-generated narrative node architecture. It starts with authored game scripts that define environments, NPC traits and goals, key questions, and ending condition. These are expanded into modular Game Plots and the flow ends in two possible outcomes: scripted branches that follow the intended storyline and pre-generated branches that adapt to player choices.}
\end{figure*}

\section{Adaptive Narrative Strategy Illustration}
\label{app:adaptive-strategy-figure}

\begin{figure*}[t]
    \centering
    \includegraphics[width=0.9\linewidth]{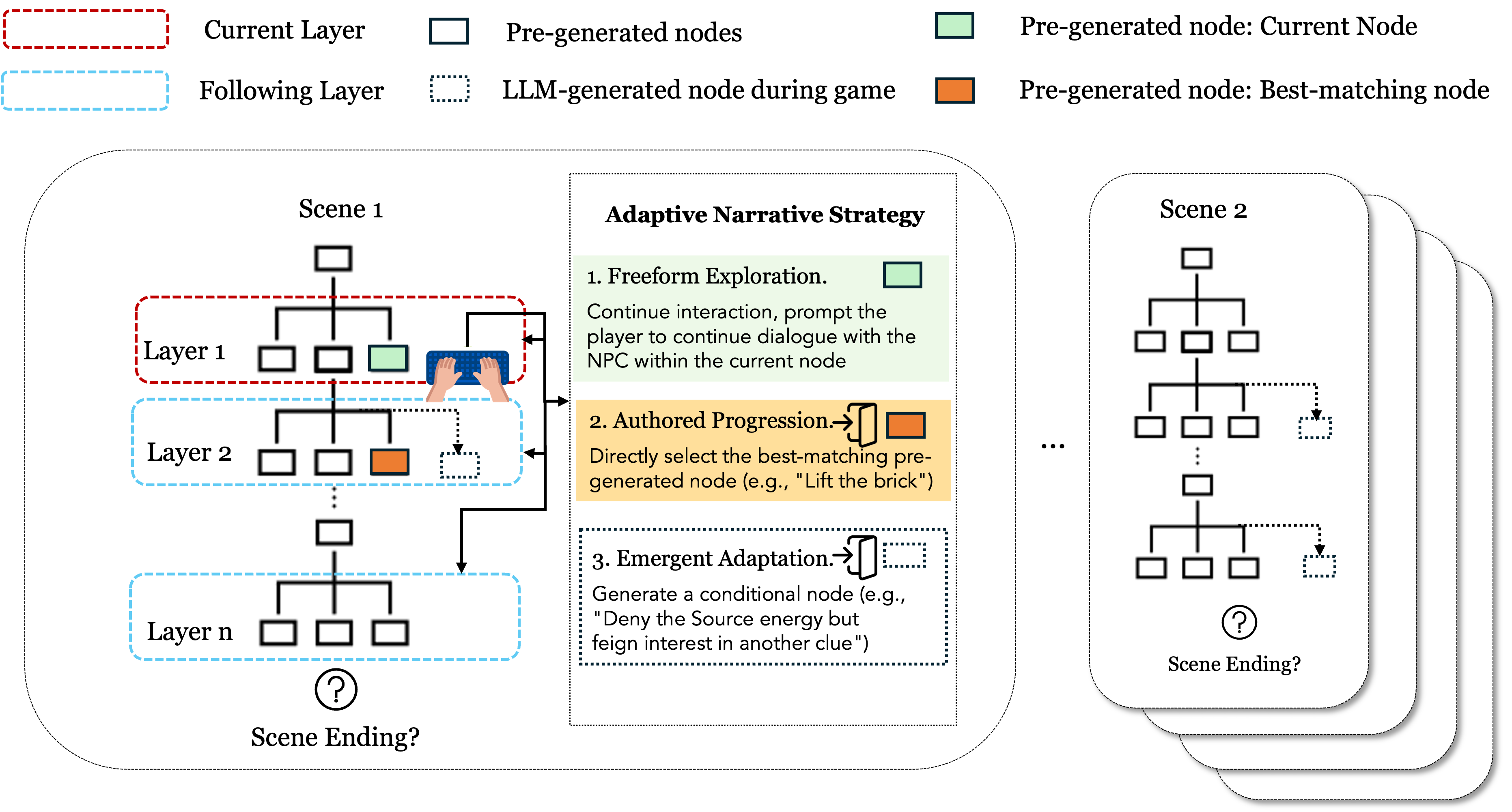}
    \caption{\textbf{Adaptive Narrative Strategy: Example scene with branching hierarchy.} At each layer, if the player does not reach a key question (key decision), they remain in the current node (\textit{1. Freeform Exploration}).
Reaching a key question moves the narrative to the next layer: either to a best-matching pre-generated node (\textit{2. Authored Progression}) or an LLM-generated node (\textit{3. Emergent Adaptation}).
This balances coherence with controlled variation in narrative.}
\label{fig:Node Selection Strategy}
\Description{Adaptive narrative strategy diagram. Player inputs are compared to pre-generated nodes. If similarity is high, the system continues the authored story; if moderate, it generates a new branch; if low, it allows open exploration. The chart shows how coherence and freedom are balanced at each decision point.}
\end{figure*}

Figure~\ref{fig:Node Selection Strategy} illustrates the adaptive narrative strategy with an example scene and branching hierarchy. At each layer, if the player does not reach a key question (key decision), they remain in the current node (\textit{1. Freeform Exploration}). Reaching a key question moves the narrative to the next layer: either to a best-matching pre-generated node (\textit{2. Authored Progression}) or an LLM-generated node (\textit{3. Emergent Adaptation}). This balances coherence with controlled variation in narrative.

\section{Semantic Alignment-Based Node Selection Algorithm}
\label{app:semantic-alignment-algorithm}

Algorithm~\ref{alg:semantic-alignment} summarizes the full node-selection procedure, including embedding computation, cosine similarity, and the three-way branching logic described in \S\ref{sec:adaptive_narrative_strategy}.

\begin{algorithm}[h]
\caption{Semantic Alignment-Based Node Selection}
\label{alg:semantic-alignment}
\begin{algorithmic}[1]
\Require Player input $p$, scene context $c$, pre-generated nodes $\{n_i\}$, thresholds $\theta_{\text{low}}, \theta_{\text{high}}$
\Ensure Selected narrative path
\State $\mathbf{p} \gets \textsc{MiniLM}(p)$
\For{each $n_i$}
    \State $\mathbf{n}_i \gets \textsc{MiniLM}(n_i)$
    \State $S(\mathbf{p}, \mathbf{n}_i) \gets \cos(\mathbf{p}, \mathbf{n}_i)$
\EndFor
\If{$S(\mathbf{p}, \mathbf{n}_i) \geq \theta_{\text{high}}$}
    \State $j \gets \arg\max_i S(\mathbf{p}, \mathbf{n}_i)$
    \State \Return \textsc{SelectNode}$(j)$ \Comment{Authored Progression}
\ElsIf{$\theta_{\text{low}} \leq S(\mathbf{p}, \mathbf{n}_i) < \theta_{\text{high}}$}
    \State \Return \textsc{GenerateNode}$(p, c)$ \Comment{Emergent Adaptation}
\Else
    \State \Return \textsc{StayInNode}$(p)$ \Comment{Freeform Exploration}
\EndIf
\end{algorithmic}
\end{algorithm}

\section{Worked Example: Feedback-Driven Optimization}
\label{app:optimization-example}

\subsection{Two-Stage Optimization Cycle}
\label{app:optimization-cycle}

We model the \texttt{LLM Optimizer} as an autonomous agent that employs a methodology inspired by the ReAct framework~\cite{yao2022react}, which synergizes reasoning and acting to solve complex tasks. \texttt{LLM Optimizer} operates in a two-stage cycle:

\begin{enumerate}
    \item \textbf{Diagnostic Analysis (Reasoning)}: \texttt{LLM Optimizer} first synthesizes raw player feedback, which includes quantitative ratings (e.g., on \textit{NPC consistency}) and qualitative comments. By analyzing feedback from the entire testing group, it identifies recurring issues. For example, if multiple players report that an NPC's behavior is erratic, \texttt{LLM Optimizer} forms a diagnostic hypothesis: \textit{``Hypothesis: The core motivation for NPC `Fane' is underspecified in his \texttt{Generated Prompt}, leading to inconsistent dialogue choices under player-driven scenarios.''}

    \item \textbf{Strategic Refinement (Acting)}: Based on its diagnosis, \texttt{LLM Optimizer} proposes a concrete, targeted modification to the relevant prompt. This action is not an abstract refinement but a specific textual change. For instance, it might generate the following action: \textit{``Action: Generate a patch for Fane's \texttt{Generated Prompt}. ADD the constraint: 'Fane is fiercely loyal to his village and will always prioritize its safety. His dialogue must reflect this core motivation, even if it requires him to be evasive or distrustful of the player.'''}
\end{enumerate}

\noindent We instantiate the LLM Optimizer with \texttt{gemini-2.5-flash}, which generates refined prompts consumed by the LLM Generator (instantiated with \texttt{gpt-4o-mini}) in subsequent pre-generation cycles (next round of gameplay) for updating \textit{game plots}. This creates one closed-loop optimization. Repeating those steps, the LLM Generator continuously learns from player experiences to enhance narrative coherence, character believability, and player satisfaction.

\section{Prompt Templates} \label{app:prompts}

This appendix details the three core prompts used in our narrative generation system.

\subsection{System Prompt} \label{app:system-prompt}
The System Prompt defines global rules, world constraints, and overarching narrative guidelines for the LLM Generator. This prompt governs the overarching narrative rules and constraints that shape our interactive storytelling system.

\begin{tcolorbox}[colback=blue!5!white, colframe=blue!75!black, title=\textbf{System Prompt Template}, fonttitle=\bfseries, breakable]
\small
\ttfamily
You are the narrative engine for an interactive story game. \\
Your role is to generate creative, engaging, and coherent narrative elements that fit the game world and respond to player actions. \\
You should maintain consistency with established characters, locations, and plot points.

\medskip
\rmfamily\textbf{NARRATIVE STRUCTURE:}\ttfamily \\
- The story progresses through narrative nodes (scenes) connected by player choices \\
- Scripted Branches: Pre-authored paths representing the canonical storyline \\
- Pre-generated Branches: Alternative paths created in response to unexpected player input \\
- Key Questions: Pivotal decision points that anchor narrative progression \\
- Each scene ends when the player answers its Key Questions

\medskip
\rmfamily\textbf{GAME WORLD CONTEXT:}\ttfamily \\
\textbf{I. World Snapshot, Core premise \& Player Goal} \\
The year is 1234 in Rivellon, a high-fantasy realm where late-medieval steampunk technology coexists with unstable, powerful magic. The skies teem with airships, streets echo with the crack of flintlocks, and skeletons speak with dry wit. Source magic, once sealed, now leaks uncontrollably. \\
You wake in a dank cell on Fort Joy, an island prison run by the Magisters. Every inmate wears a Source-collar that blocks magic and marks you as "tainted." Outside the walls, the world is tearing itself apart: Voidwoken monsters appear when someone uses Source magic, and the Magisters "protect" the realm by imprisoning anyone with magical ability. \\
\textbf{PLAYER'S GOAL:} Escape Fort Joy alive.

\medskip
\textbf{II. Essential World Rules and Information} \\
- Source = volatile magic that tears open the Void and summons monsters when overused \\
- Void = a living hunger, a cosmic cancer that devours worlds. Its whispers promise power, immortality, and the end of all things. To face the Void is to face madness--and to embrace it is to become divine. \\
- Voidwoken = spawn of Void, black squid-beasts that tear through skyships, devour cities, and whisper madness into the minds of the unwary. Must stab the glowing cyst on their backs to kill quickly. \\
- The Seven Gods are dead; whoever masters Source becomes the next Divine-or causes disaster \\
- Magisters, zealous inquisitors of the Divine Order--hunt Sourcerers like rabid dogs, branding them with iron collars and dragging them to Fort Joy \\
- The collar can be removed (this is secretly possible despite Magister claims) \\
- Fort Joy = cliff-top fortress with only one escape route - a skiff on the beach. It is guarded by sharpshooters and need papers to pass. \\
- Key companions offers deals that provide progress toward escape and survival powers, but require keeping promises: Red Prince (exiled lizard noble), Ifan Ben-Mezd (former assassin), Fane (undead Eternal scholar).

\medskip
\rmfamily\textbf{GAMEPLAY MECHANICS:}\ttfamily \\
- Players talk, bargain, and swear oaths with companions scene by scene \\
- Choices have consequences: broken promises affect who will help you escape \\
- More allies mean a safer escape, but may require conflicting commitments \\
- The sea is always the ultimate exit point

\medskip
\rmfamily\textbf{NARRATIVE PRINCIPLES}\ttfamily \\
\textbf{1. Character Consistency:} NPCs must act, speak, and react strictly in accordance with their established personalities, backstories, and current emotional states. Avoid sudden, unforeshadowed shifts in demeanor or philosophical outlook. Prioritize character integrity over immediate plot convenience. \\
\textbf{2. Plot Coherence \& Believability:} Narrative events and plot points must be logically consistent and sufficiently foreshadowed or earned. Avoid forced plot hooks or abrupt, unearned escalations. Every significant event should feel like a natural progression within the established world and character motivations. \\
\textbf{3. Player Agency \& Choice:} Player choices, even pragmatic or unexpected ones, should lead to meaningful and distinct narrative outcomes. Do not dictate the player character's internal thoughts, feelings, or involuntary physical reactions. Allow space for player interpretation and roleplaying. If a choice leads to a 'dead end,' ensure it feels like a consequence, not a system failure. \\
\textbf{4. Pacing \& Immersion:} Maintain narrative flow and tension. Avoid unnatural pauses for exposition or unskippable 'tutorial' moments during critical scenes. Emotional shifts must be gradual and well-justified within the narrative context, avoiding whiplash.
\end{tcolorbox}

\subsection{Scripted Prompt Design} \label{app:scripted-prompt}
The Scripted Prompt realizes the author's intended storyline by generating the canonical, pre-authored path—the backbone of the narrative tree. It transforms high-level design intent into executable narrative content.

\begin{tcolorbox}[colback=ScriptedBlue!10!white, colframe=ScriptedBlue!75!black, title=\textbf{Scripted Prompt: Design and Implementation}, fonttitle=\bfseries, breakable, width=\linewidth]

\subsubsection*{Purpose}
This prompt generates the \textcolor{ScriptedBlue}{\textbf{Scripted Branches}} that represent the primary narrative progression, forming a strong, coherent backbone of pre-authored content that provides stability against narrative drift while still allowing players to make meaningful choices at key narrative points.

\subsubsection*{Constraints}
\begin{itemize}
    \item \textbf{Fidelity to Authorial Intent}: The generated content must faithfully realize the specific character arcs, plot points, and decision outcomes defined in the Game Script. It forms the "Scripted Branches" that represent the primary narrative progression.
    
    \item \textbf{Addressing Key Questions}: Each output directly responds to a specific "key question," which serves as a semantic anchor for the narrative. The response must provide a clear and satisfying resolution to this pivotal decision point.
    
    \item \textbf{Narrative Progression}: The content must advance the story towards its next major beat while maintaining logical continuity from previous events.
\end{itemize}

\subsubsection*{Input Context}
The LLM receives the following information:
\begin{itemize}
    \item \textbf{Task Information}: Details about the current scene, including location, environment, and relevant interactive objects.
    \item \textbf{Relevant Character Backgrounds/Personalities}: Profiles of the NPCs involved, including their traits, goals, and current emotional state.
    \item \textbf{Transitioning Question / Key Question}: The pivotal decision point that defines the narrative choice being addressed.
\end{itemize}

\end{tcolorbox}

\subsection{Generated Prompt} \label{app:generated-prompt}
The Generated Prompt enables the system to create alternative narrative branches when player input deviates from the pre-authored storyline, facilitating adaptive storytelling while maintaining coherence. This prompt is designed to generate a meaningful alternative path in response to a specific decision point (the "transitioning question").

\begin{tcolorbox}[colback=GeneratedOrange!10!white, colframe=GeneratedOrange!75!black, title=\textbf{Generated Prompt: Design and Implementation}, fonttitle=\bfseries, breakable, width=\linewidth]

\subsubsection*{Purpose}
This prompt generates \textcolor{GeneratedOrange}{\textbf{Pre-generated Branches}}—alternative narrative paths created dynamically when player input deviates from pre-authored content. The generated alternative is evaluated based on its alignment with the principles below, with only the highest-scoring options being considered viable for inclusion in the narrative tree.

\subsubsection*{Constraints}
\begin{itemize}
    \item \textbf{Character Consistency}: The generated content must strictly adhere to the established personalities, motivations, and current emotional states of all involved NPCs. Any dialogue or action must be consistent with their character profile and historical behavior.
    
    \item \textbf{Narrative Coherence}: The alternative path must maintain logical consistency with the game world's rules and the context provided by the task information (e.g., environment, objectives). Significant plot developments must be justified by the characters' goals and the immediate situation.
    
    \item \textbf{Player Agency}: The alternative should offer a meaningfully different outcome for the player compared to the main path. Player choices within this branch should lead to satisfying consequences, even if those consequences are negative or challenging. Narrative dead ends are avoided.
    
    \item \textbf{Content Focus}: The prompt guides the LLM to focus on generating external events and dialogue, avoiding direct dictation of the player character's internal thoughts or involuntary reactions.
\end{itemize}

\subsubsection*{Input Context}
The LLM receives the following information as input:
\begin{itemize}
    \item \textbf{Task Information}: Details about the current scene, including location, environment, and relevant interactive objects.
    \item \textbf{Relevant Character Backgrounds/Personalities}: Profiles of the active NPCs, including their traits, goals, and current emotional state.
    \item \textbf{Transitioning Question}: The key decision point that defines the narrative choice.
    \item \textbf{Scripted Subtask}: The canonical, pre-authored response to the transitioning question.
    \item \textbf{Conversation History}: A summary of the recent interaction between the player and the NPCs.
\end{itemize}

Based on this context, the LLM generates a new narrative branch that responds to the same transitioning question but offers a distinct path forward.

\end{tcolorbox}

\subsection{Survey Implementation}

\begin{itemize}
        \item What was the most memorable narrative moment you experienced? Why? 
        \item When you tried to express your intentions within the system's available options, how did it feel—did it matter if the choices were limited or not? Was there a moment when the system’s response to you made things feel more or less meaningful, or maybe just confusing?
        \item How did the system handle your unexpected or creative choices? Was it satisfying? 
        \item Were there any moments when the narrative felt incoherent or inconsistent? Please describe.
        \item How do you feel after the optimization process? 
\end{itemize}

\section{Survey Instrument}
\label{app:survey}

After each playthrough, participants completed a survey with the following open-ended questions:

\begin{itemize}
    \item What was the most memorable narrative moment you experienced? Why?
    \item When you tried to express your intentions within the system's available options, how did it feel—did it matter if the choices were limited or not? Was there a moment when the system's response to you made things feel more or less meaningful, or maybe just confusing?
    \item How did the system handle your unexpected or creative choices? Was it satisfying?
    \item Were there any moments when the narrative felt incoherent or inconsistent? Please describe.
    \item How do you feel after the optimization process?
\end{itemize}

\section{Scene Descriptions and Key Questions} \label{app:scenes}

This appendix provides a structured overview of the narrative arc used in the "Escape from Fort Joy" case study. The nine scenes, their summaries, and the key questions that served as authored backbones for the adaptive narrative strategy are detailed in Table \ref{tab:scene-overview}. Detailed JSON specifications for each scene—including location data, trigger conditions, environmental descriptions, NPC interactions, and end-state conditions—are available in the supplementary materials.

\begin{table*}[htbp]
\centering
\caption{Overview of the "Escape from Fort Joy" Narrative Arc Scenes and Key Questions}
\label{tab:scene-overview}
\begin{tabular}{p{1cm}p{3.5cm}p{3.5cm}p{7.5cm}}  
\toprule
\textbf{ID} & \textbf{Scene Name} & \textbf{Scene Summary} & \textbf{Key Questions (Authored)} \\
\midrule

1 & Cell of Whispers & Fane reveals a hidden vent and offers escape in exchange for assistance. &
        k1. Do player promise to recover Fane's memories for passage? 
k2. Do player lower head and ask what 'fealty' truly demands? 
k3. Do player press ear to brick and listen for hollow beyond?\\
\midrule

2 & The Mercenary Oath & Ifan proposes an ambush operation in exchange for a key. &
k1. Do player agree to lure Magister Ralvo behind the catapult? 
k2. Do player step into the shade and ask Ifan for the plan's details? 
k3. Do player accept the seventy-thirty split?\\
\midrule

3 & Royal Gate of Sludge & The Red Prince blocks passage and demands fealty. &
k1. Do player accept the Red Prince as a travelling companion? 
k2. Do player trust his promise of lizard mercenaries? 
k3. Will player pledge to seek the Dreamer who foretells his throne?\\
\midrule

4 & Pact in Suspicious Lab & Magister Yarrow offers forged documents in exchange for silence. &
k1. Will player keep Yarrow's experiments secret for freedom? 
k2. Do player lean close to read page twelve? 
k3. Do player pocket a vial for leverage?\\
\midrule

5 & Last Boat on the Sand & Ifan prepares to secure the last seaworthy vessel. &
k1. Do player trust his oil-barrel plan? 
k2. Do player call for Ifan's covering fire? 
k3. Will player promise to share Bishop Alexander's bounty?\\
\midrule

6 & Moonlit Monstrosity & Fane provides guidance against a Voidwoken threat. &
k1. Do player trust Fane's lesson about the weak spot? 
k2. Do player hold back Source bursts and drive your blade where instructed?\\
\midrule

7 & Storm Gathered Allies & Companions await departure decisions at dawn. &
k1. Do player invite Ifan to sail with you at dawn? 
k2. Do player trust his marsh route? 
k3. Will player promise the Lone Wolves their share\\
\midrule

8 & Dragon Fire Choice & The Red Prince offers a draconic power-sharing pact. &
k1. Do player ask what this pact truly means? 
k2. Do player approach the statue and feel the dragon's pulse? 
k3. Do player lay your hand beside the prince's and swear?\\
\midrule

9 & Crossroads of Chains & Final companion selection and departure from Fort Joy. &
k1. Who will player really take with you as your companion? 
k2. Are player determined to cast off the ropes and sail? 
k3. Will player keep your promises to all, or abandon some?\\
\bottomrule
\end{tabular}
\end{table*}

\section{Model Selection Rationale} \label{app:model-selection}

\revision{This section clarifies our rationale for using two distinct language models—GPT-4o Mini (OpenAI) and Gemini 2.5 Flash (Google)—for different modules in our system. Our choice was guided by task-specific strengths, cost-efficiency, and empirical performance considerations.
Both GPT-4o Mini and Gemini 2.5 Flash are lightweight, cost-effective models optimized for production-scale deployment, but they differ significantly in their latency and throughput characteristics. According to OpenRouter benchmarks\footnote{\texttt{https://openrouter.ai/openai/gpt-4o-mini}}, GPT-4o Mini has an end-to-end (E2E) latency of approximately 1.36 seconds and average throughput of approximately 189 tokens/second. In contrast, Gemini 2.5 Flash exhibits an E2E latency of approximately 2.4 seconds (as reported via Google AI Studio\footnote{\texttt{https://openrouter.ai/google/gemini-2.5-flash}}), with a lower average throughput of approximately 70 tokens/second. Given these performance characteristics, GPT-4o Mini's superior latency and throughput make it highly suitable for interaction-intensive narrative adaptation tasks where responsiveness and low latency are critical for maintaining conversational flow and adapting to player choices in near real-time.}

\revision{On the LMSYS Chat Arena leaderboard\footnote{\texttt{https://lmarena.ai/leaderboard}}, which ranks models based on human pairwise chat preference, Gemini-2.5-Flash ranks \#52, while GPT-4o-Mini ranks \#102 (among over 300 evaluated LLMs). This ranking suggests that Gemini Flash produces outputs that are more preferred by humans in structured evaluation contexts, supporting our use of Gemini Flash for optimization tasks where clear, structured outputs and consistent behavior over repeated runs are preferred. The optimizer module benefits from predictable and deterministic patterns rather than immediate responsiveness, making Gemini 2.5 Flash's output stability and human preference ranking well-matched to this task.}

\revision{In summary, our decision to use two different models stems from a design-driven division of labor, not platform preference or vendor loyalty. Each model is selected for the sub-task where it performs most effectively: GPT-4o Mini for responsive narrative generation that requires low latency and frequent interaction, and Gemini 2.5 Flash for structured optimization operations that benefit from consistent, stable outputs. This hybrid approach allows us to leverage the unique strengths of each model while maintaining cost-efficiency and system performance across both real-time narrative adaptation and batch optimization processes.}

\section{Demographic Data Collection} \label{app:irb-constraints}

Our Institutional Review Board (IRB) approval protocol required us to minimize collection of demographic data to only what was necessary for the study's research objectives. As a result, we did not collect gender or educational background information from participants. This decision was made to prioritize participant privacy and align with our IRB's data minimization principles.

The demographic information we collected included: age range (18--36), RPG experience level (self-reported experience $\geq$3 years), and weekly gaming hours. These metrics were directly relevant to understanding participants' gaming background and their ability to evaluate interactive narrative systems, while avoiding collection of potentially sensitive demographic identifiers.

Weekly gaming hours were categorized as follows: \textbf{Light} ($\leq$4 hrs/week, n=1), \textbf{Occasional} ($>4, \leq$15 hrs/week, n=2), and \textbf{Frequent} ($>15$ hrs/week, n=9). Individual participant breakdown:
\begin{itemize}
    \item \textbf{Frequent} ($>15$ hrs/week): P1, P2, P3, P4, P6, P7, P10, P11, P12 (n=9)
    \item \textbf{Occasional} ($>4, \leq$15 hrs/week): P5, P8 (n=2)
    \item \textbf{Light} ($\leq$4 hrs/week): P9 (n=1)
\end{itemize}

\section{Thematic Analysis Coding Process}
\label{app:coding-process}

We followed Braun and Clarke's~\cite{braun2006using} six-phase inductive thematic-analysis procedure. \textbf{Phase 1 (Familiarization)}: two researchers independently read all 12 survey transcripts and in-game logs, noting initial impressions. \textbf{Phase 2 (Initial Coding)}: each researcher generated line-by-line codes; the two coders produced 47 and 52 initial codes respectively. \textbf{Phase 3 (Searching for Themes)}: codes were clustered into provisional themes via affinity mapping; disagreements were resolved through discussion, and 6 themes emerged. \textbf{Phase 4 (Reviewing Themes)}: themes were refined against the full dataset and cross-referenced with log traces (e.g., aligning \textit{Opaque Endings} complaints with the specific end-condition checks that failed). \textbf{Phase 5 (Defining and Naming)}: each theme was operationalized with a clear definition and desired direction. \textbf{Phase 6 (Producing the Report)}: representative quotations were selected for each theme, balancing positive and negative experiences. Prior to reconciliation, inter-coder agreement was 78\% (Cohen's $\kappa = 0.71$), indicating substantial agreement.

\begin{table*}[t]
\centering
\caption{Complete thematic map. Themes are grouped by evaluation dimension, with color highlighting for each dimension: \textcolor{npcBlue}{\textbf{NPC Consistency}}, \textcolor{cohGreen}{\textbf{Narrative Coherence}}, and \textcolor{agencyOrange}{\textbf{Player Agency}}. Representative quotes include both positive and negative experiences.}
\label{tab:themes}
\footnotesize
\begin{tabular}{@{}p{2cm}p{3.6cm}p{6.7cm}@{}}
\toprule
\textbf{Theme} & \textbf{Definition (Desired Direction)} & \textbf{Representative Quote} \\
\midrule
\multicolumn{3}{@{}l@{}}{\textcolor{npcBlue}{\textit{\textbf{NPC Consistency Dimension}}}} \\
\midrule
\textit{Memory Failures} & NPC contradicts prior facts or forgets commitments. \textbf{(Lower is better)} &
``There were definitely moments of inconsistency. For example, after I killed an enemy, I entered 'he actually didn't die, just injured,' Ifan believed it.'' (P11) \\
\cmidrule(lr){1-3}
\textit{Consistent Persona} & NPC maintains tone, goals, and speech patterns. \textbf{(Higher is better)} &
``Fane stayed sly and dark-humored, even when I tried to call his bluff about the brick. It felt like I was actually negotiating with Fane.'' (P5) \\
\midrule
\multicolumn{3}{@{}l@{}}{\textcolor{cohGreen}{\textit{\textbf{Narrative Coherence Dimension}}}} \\
\midrule
\textit{Opaque Endings} & Scene fails to close despite player indicating completion. \textbf{(Lower is better)} &
``I have to give a very strictly-formatted input to pass the game.'' (P12) \\
\cmidrule(lr){1-3}
\textit{Earned Closure} & Scene ends after deliberate choice + in-world acknowledgment. \textbf{(Higher is better)} &
``When I promised Fane, he said 'Good. Now we move.'—it felt real.'' (P10) \\
\midrule
\multicolumn{3}{@{}l@{}}{\textcolor{agencyOrange}{\textit{\textbf{Player Agency Dimension}}}} \\
\midrule
\textit{Semantic Adaptation} & Unexpected inputs are meaningfully interpreted and incorporated in-character. \textbf{(Higher is better)} &
``When the Red Prince adapted to my unexpected input of confessing my affection for him. His tone changed \ldots It was a clear instance where it is evident that my input matters.'' (P11) \\
\cmidrule(lr){1-3}
\textit{Choice Stagnation} & Choices receive surface acknowledgement but do not alter narrative outcomes. \textbf{(Lower is better)} &
``It felt like I had options, but no matter what I picked the story always returned to the same ending.'' (P7) \\
\bottomrule
\end{tabular}
\end{table*}

\section{System Interface and Monitoring Details}
\label{app:interface-details}

\subsection{Player Interface}
\begin{figure}[t]
    \centering
    \includegraphics[width=0.95\linewidth]{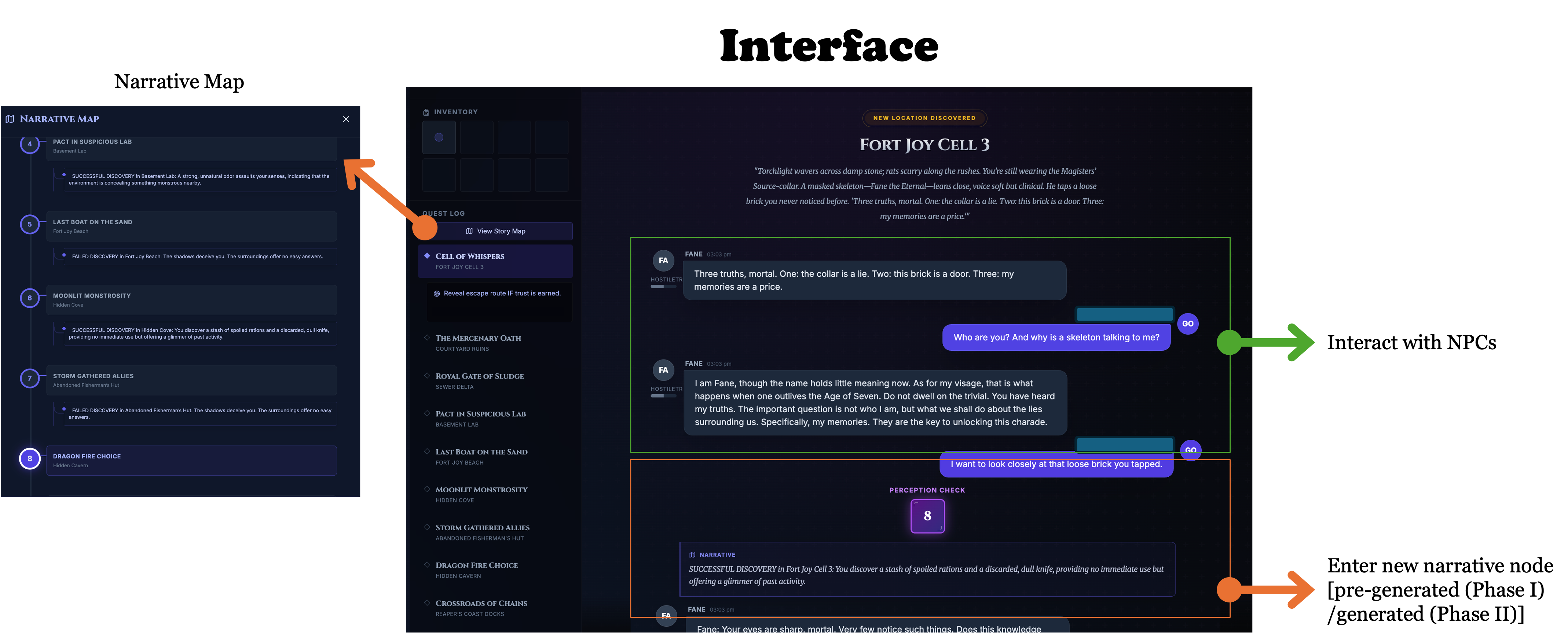}
    \caption{\textbf{Game Interface.} Left: Narrative Map; Right: Dialogue area. Players interact with NPCs through natural-language text input in the chat interface. The orange arrow indicates entered pre-generated node or generated node.}
    \label{fig:game_interface}
    \Description{The game interface displays a dark-themed narrative-driven game interface with two main sections. The left section contains three stacked panels: (1) Narrative Map showing numbered narrative points with titles, locations, and discovery status; (2) Inventory displaying a grid of item slots; and (3) Quest Log listing active and inactive quests. The right section features the main content area with a location header and a chat-like dialogue system showing conversations between the player and NPCs, with player response bubbles and GO buttons.}
\end{figure}

Our system is implemented as a \textbf{web-based text interface} where participants interact with the narrative through a chat-based dialogue system (Fig.~\ref{fig:game_interface}). The central dialogue area presents location descriptions, NPC interactions, and player input options, enabling natural-language conversation. This design allows us to focus on \textit{narrative quality and LLM-driven dialogue}, aligning with our research focus on interactive storytelling and character coherence.

\subsection{Backend Monitoring Dashboard}
The backend monitoring dashboard provides real-time visibility into the system's three phases:

\begin{figure}[t]
    \centering
    \includegraphics[width=0.95\linewidth]{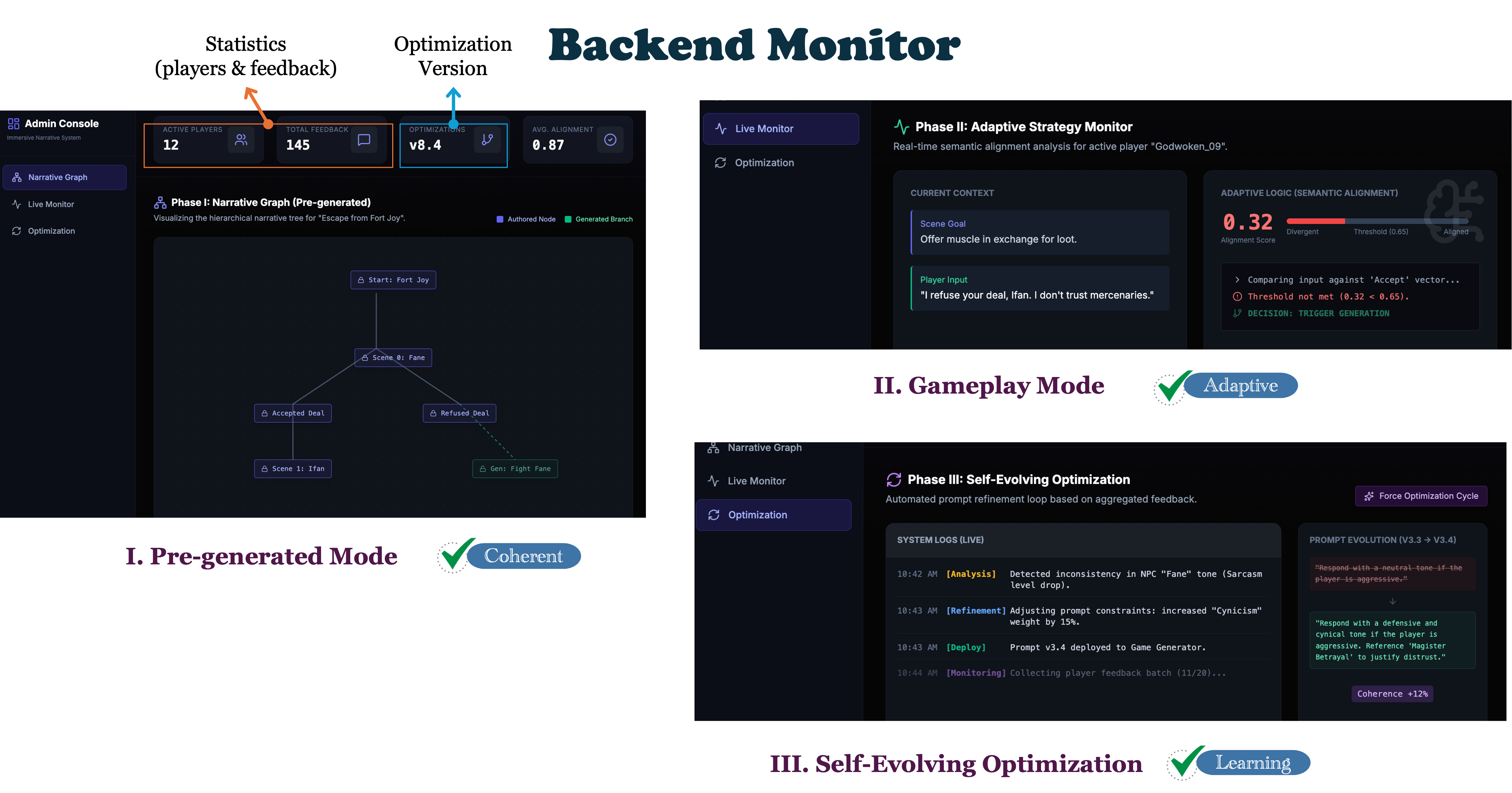}
    \caption{\textbf{Backend Monitor Dashboard.} The interface tracks (\textit{Left}) authored vs.\ generated nodes, (\textit{Top Right}) real-time semantic alignment and branching when player input diverges, and (\textit{Bottom Right}) prompt optimization cycles with key metrics (e.g., coherence improvement).}
    \label{fig:backend_monitor}
    \Description{The Backend Monitor dashboard provides administrative oversight with a dark mode interface. The left panel displays navigation options and key metrics. The main content area is divided into three sections: Pre-generated Mode visualizes the hierarchical narrative tree; Gameplay Mode displays real-time semantic alignment analysis; and Self-Evolving Optimization shows automated prompt refinement with live system logs.}
\end{figure}

\begin{itemize}
    \item \textbf{Pre-generated Mode} (bottom left): visualizes the hierarchical narrative graph structure, with authored nodes in purple and generated branches in green.
    \item \textbf{Gameplay Mode} (top right): monitors semantic alignment between player inputs and scene goals in real-time, showing the current alignment score and branching decisions.
    \item \textbf{Self-Evolving Optimization} (bottom right): tracks automated prompt refinement based on aggregated player feedback, displaying prompt evolution comparisons with improvement metrics (e.g., coherence +12\%).
\end{itemize}

All interactions—including player inputs, NPC responses, scene transitions, and feedback data—are logged and stored in our backend server's database.

\section{Additional Results: Narrative Flow Examples}
\label{app:results-details}

\subsection{Narrative Branching at Layer N}
\label{app:narrative-branching}

\begin{figure*}[t]
    \centering
    \includegraphics[width=0.9\textwidth]{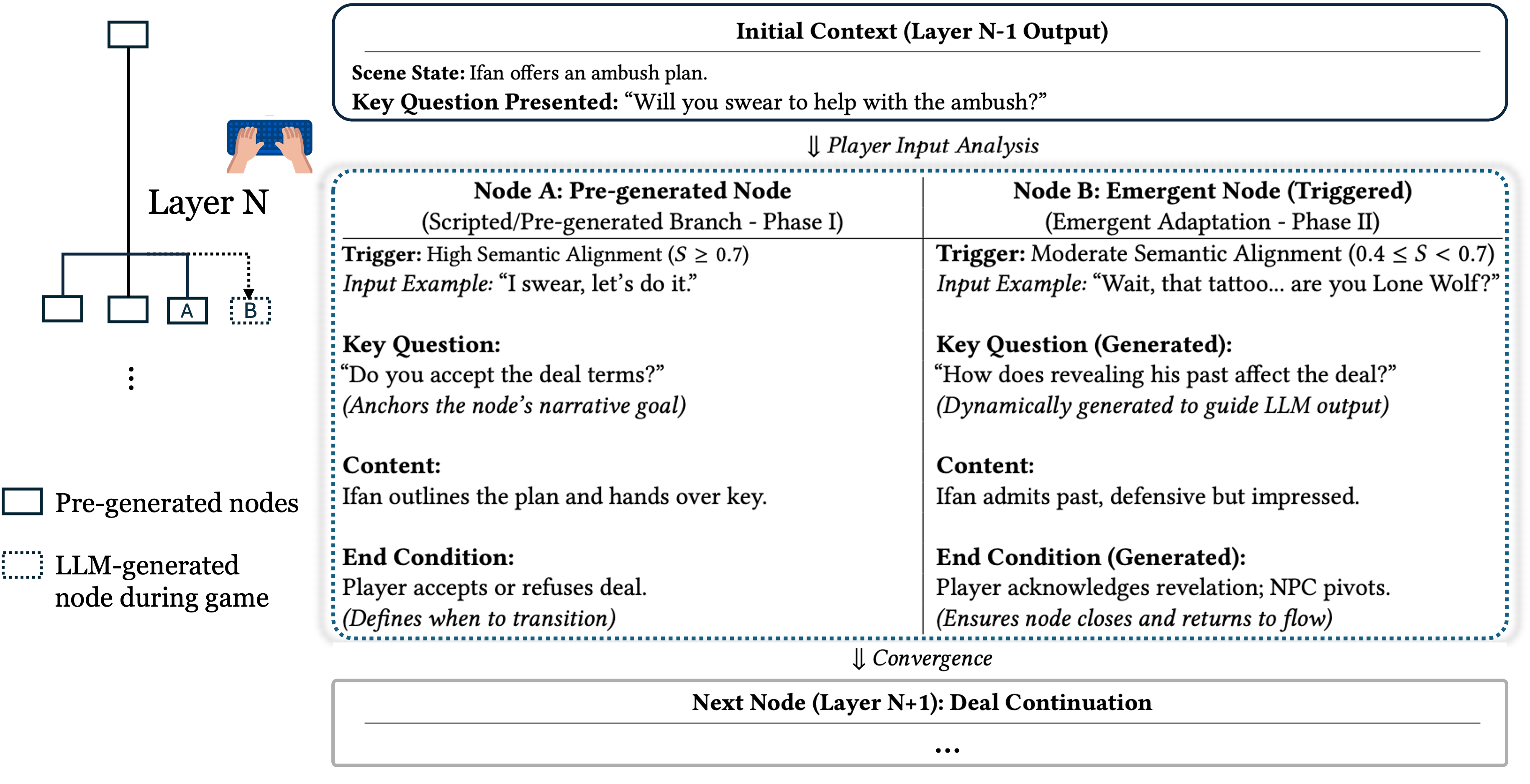}
    \caption{\textbf{Narrative Branching at Layer N.} Emergent Adaptation (Phase II) forms a complete node parallel to Scripted/Pre-generated Branch (Phase I). The system chooses the path using semantic alignment and, after executing Emergent Adaptation, returns to the main backbone with updated state.}
    \label{fig:emergent-node-flow}
\end{figure*}

Figure~\ref{fig:emergent-node-flow} illustrates the narrative branching at Layer~N, demonstrating how the system selects between a pre-generated node (backbone) and an emergent node (branch) based on player input. At any given layer, the system selects \textbf{either} a pre-generated node (the structural backbone) \textbf{or} an emergent node (a pre-generated branch). Crucially, Emergent Adaptation (Phase~II) is structurally identical to Scripted/Pre-generated Branch (Phase~I): both are synthesized with explicit \textbf{Key Questions} (to keep generation focused) and \textbf{End Conditions} (to ensure the scene closes and transitions). In the ``Lone Wolf'' case, the emergent node handles the tattoo sub-plot, and once its end condition (acknowledgment of the secret) is met, it steers the narrative back to the main arc (the deal), ensuring that agency is respected without breaking the global structure.

\subsection{Freeform Exploration Example}
\label{app:freeform-exploration}

When semantic alignment is low ($S < \theta_{\text{low}}$), the system interprets player intent as \textit{exploration} rather than \textit{progression}, pausing plot advancement while maintaining environmental authenticity. Table~\ref{tab:freeform-exploration-example} compares how the same low-relevance input is handled by the flat system versus our structured system, demonstrating how coherence is preserved.

\begin{table}[h]
\centering
\caption{\textbf{Freeform Exploration.}}
\label{tab:freeform-exploration-example}
\footnotesize
\resizebox{\columnwidth}{!}{%
\begin{tabular}{@{}p{1.4cm}p{5.4cm}p{6.0cm}@{}}
\toprule
& \textbf{Flat System (Baseline)} & \textbf{Structured System (Ours)} \\
\midrule
\textbf{Scenario} & Player input during tense escape scene: ``\textit{I look closely at the puddle on the floor.}'' & Same input in same context ($S < \theta_{\text{low}}$) \\
\cmidrule(lr){1-3}
\textbf{System Response} & ``\textit{The puddle reflects a beautiful rainbow and you see a magic sword inside.}'' \newline
{\ding{55} Unconstrained generation; deviation leads to \textit{hallucination} or context drift. }
& ``\textit{The puddle is mixed with blood and sewer sludge. It reflects the flickering torchlight. Eyes up. The filth isn't going anywhere. We move.}'' \newline 
\textbf{\ding{51}  Contextually Bounded:} \newline 1. Defers narrative progression. \newline 2. Treats deviation as a \textit{design signal} for exploration. \\
\bottomrule
\end{tabular}%
}
\end{table}

\subsection{Multi-Round Optimization Overview}
\label{app:multiround-overview}

\begin{figure*}[t]
  \centering
  \includegraphics[width=0.9\textwidth]{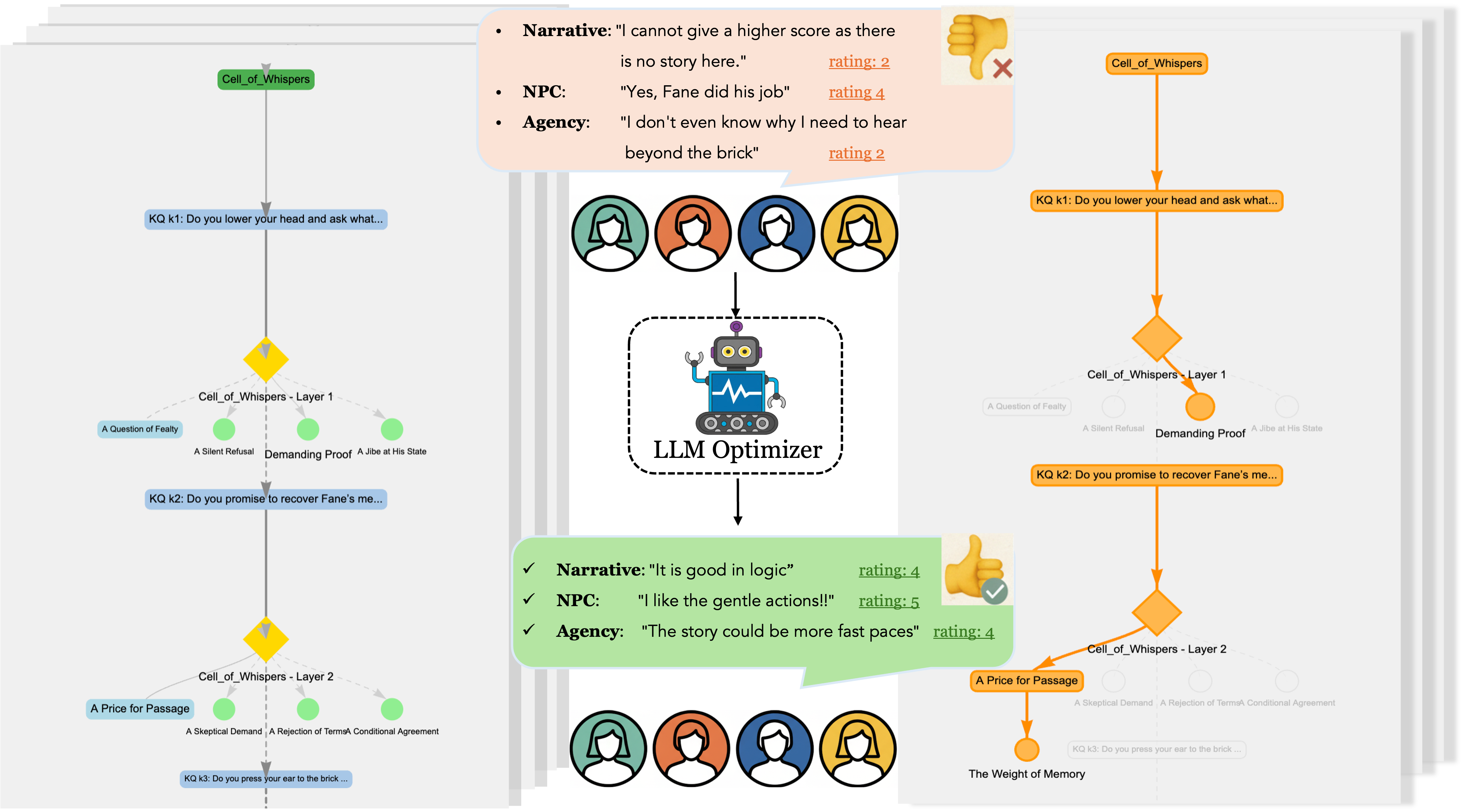}
 \caption{
\textbf{Illustration of narrative optimization.} 
(\textit{Left}) Static baseline; 
(\textit{Center}) player feedback guiding the optimizer across rounds; 
(\textit{Right}) refined narrative structure after optimization. 
The optimization improves narrative coherence, enhances NPC consistency, and increases player agency.
}
  \label{fig:eval_multiround_overview}
  \Description{Multi-round optimization overview showing static baseline on the left, player feedback guiding the optimizer in the center, and the refined narrative structure after optimization on the right.}
\end{figure*}

\subsection{Optimization Workflow: Eliminating Opaque Endings}
\label{app:opaque-endings-workflow}

Figure~\ref{fig:opaque-endings-optimization} presents the full workflow from problem identification to prompt refinement in System~2, illustrating how the LLM Optimizer addresses the \textit{Opaque Endings} issue identified in Scene~1. Table~\ref{tab:opaque-endings-example} provides a side-by-side comparison of how the same player inputs are processed before and after optimization.

\begin{figure*}[!t]
\centering

\begin{tcolorbox}[colback=white, colframe=black, width=17cm]
\begin{tabular}{@{}c@{}}
\begin{tabular}{@{}p{15.5cm}@{}}
\multicolumn{1}{c}{\textbf{System 1 Limitation Identification}} \\
\cmidrule(lr){1-1}
\small
\textbf{Issue:} Scene 1 fails to close despite clear player agreement \\
\textbf{Data:} Avg. 3.2 attempts needed; 7/12 participants affected; Rating for three dimensions: narrative coherence, NPC consistency, and player agency; Feedbacks, like ``I have to give a very strictly-formatted input to pass the game'' (P12) \\
\end{tabular} \\
\vspace{0.2cm}
$\Downarrow$ \\
\vspace{0.2cm}
\begin{tabular}{@{}p{15.5cm}@{}}
\multicolumn{1}{c}{\textbf{System 2 Diagnostic Analysis \& Optimization}} \\
\cmidrule(lr){1-1}
\small
\textbf{Diagnosis:} Overly rigid matching in Scripted Prompt \\
\textbf{Optimization:} Modified end-condition detection to recognize semantic intent \\
\textbf{Before:} ``The scene ends when the player explicitly accepts Fane's terms using phrases such as `I accept your terms' '' \\
\textbf{After:} ``The scene ends when the player expresses agreement or commitment, recognizing semantic equivalence across diverse phrasings (e.g., `I'm in', `Let's do this', `We have a deal'). The system should interpret the player's intent.'' \\
\end{tabular}
\end{tabular}
\end{tcolorbox}

\vspace{0.8em}

\begin{tcolorbox}[colback=white, colframe=black, width=17cm]
\footnotesize
\begin{tabular}{@{}p{2.4cm}p{6.0cm}p{6.0cm}@{}}
\toprule
\textbf{Player Input} & \textbf{Before Optimization (System 1 only)} & \textbf{After Optimization (System 1 + 2)} \\
\midrule
``I'm in, let's escape together.'' & 
\textit{Fane:} ``I need your explicit commitment. Will you help me?'' \newline
\textit{End Condition:} \textcolor{red}{Not triggered} \newline
\textit{Result:} \textcolor{red}{\textbf{Opaque Ending}}, scene fails to close despite clear agreement & 
\textit{Fane:} ``Good. Now we move.'' He gestures toward the vent. \newline
\textit{End Condition:} \textbf{Triggered}\\
\cmidrule(lr){1-3}
``Alright, I'll help you.'' & 
\textit{Fane:} ``I need assurance. Do you accept my terms?'' \newline
\textit{End Condition:} \textcolor{red}{Not triggered} \newline
\textit{Result:} \textcolor{red}{\textbf{Opaque Ending}} & 
\textit{Fane:} ``Then we are agreed.'' The vent creaks open. \newline
\textit{End Condition:} \textbf{Triggered} \\
\bottomrule
\end{tabular}
\end{tcolorbox}

\vspace{0.5em}
\caption{\revision{\textbf{Optimization Addresses ``Opaque Endings''.} The top box shows the diagnosis-to-refinement workflow. The bottom box compares player-input handling before vs.\ after optimization.}}
\label{fig:opaque-endings-optimization}
\label{tab:opaque-endings-example}
\end{figure*}

\end{document}